%% file: neurips_2024.tex
\documentclass{article}

\PassOptionsToPackage{table,xcdraw,dvipsnames}{xcolor}

\usepackage[final]{neurips_2024}

\usepackage[utf8]{inputenc} % allow utf-8 input
\usepackage[T1]{fontenc}    % use 8-bit T1 fonts
\usepackage{hyperref}       % hyperlinks
\usepackage{url}            % simple URL typesetting
\usepackage{booktabs}       % professional-quality tables
\usepackage{amsfonts}       % blackboard math symbols
\usepackage{nicefrac}       % compact symbols for 1/2, etc.
\usepackage{microtype}      % microtypography
\usepackage[dvipsnames]{xcolor}         % colors
\usepackage{titlesec}
\usepackage{ulem}
\usepackage[font=small]{caption}
\usepackage{layout} % 用于输出页面布局信息

\usepackage{tcolorbox}

\titlespacing{\paragraph}{%
0pt}{%
0.01 \baselineskip}{%
1em}%

\definecolor{darkorange}{RGB}{255, 140, 0}
\definecolor{darkblue}{RGB}{84, 112, 198}
\definecolor{lightgreen}{RGB}{145, 204, 117}
\definecolor{lightyellow}{RGB}{250, 200, 88}
\definecolor{lightred}{RGB}{238, 102, 102}
\definecolor{lightblue}{RGB}{115, 192, 222}
\newtcolorbox{promptbox}[2][Prompt]{
  colback=black!5!white,
  arc=5pt,
  boxrule=0.5pt,
  fonttitle=\bfseries,
  title=#1,
  before upper={\small}, fontupper=\fontfamily{ptm}\selectfont,
  colframe=#2, % 使用传递的参数来设定 colframe
}
\usepackage[utf8]{inputenc}
\usepackage{newunicodechar}

\newunicodechar{─}{\rule[0.5ex]{1em}{0.2ex}}
\newunicodechar{│}{\textbar}
\usepackage{verbatim}
\usepackage{microtype}
\usepackage{hyperref}
\usepackage{url}
\usepackage{booktabs}
\usepackage{enumitem}
\usepackage{graphicx}
\usepackage{multirow}
\usepackage{pifont}
\usepackage{booktabs}
\usepackage{hyperref}
\usepackage{makecell}
\usepackage{amsmath}
\usepackage{amsfonts}
\usepackage{enumitem}
\usepackage{multicol}
\usepackage{soul}
\usepackage{arydshln}
\usepackage{subcaption}
\usepackage{wrapfig}
\usepackage{xspace}

\usepackage{caption}
\usepackage{color}  % red text

\usepackage{todonotes}

\usepackage{algorithm}
\usepackage{algpseudocode}
\usepackage{threeparttable}

\input{math_commands}

\newcommand{\rawmethod}[1][]{DART#1}
\newcommand{\method}[1][]{\texttt{DART#1}}
\newcommand{\fullmethod}{Difficulty-Aware Rejection Tuning}
\newcommand{\model}[1][]{\texttt{\method-Math#1}}
\newcommand{\dataset}[1][]{\texttt{\method-Math#1}}
\newcommand{\sampling}[1][]{\texttt{DARS#1}}

\newcommand{\darsu}{\sampling[-Uniform]}
\newcommand{\darsp}{\sampling[-Prop2Diff]}

\newcommand{\up}[1]{\textcolor{OliveGreen}{\small \ $\uparrow${#1}}}
\newcommand{\down}[1]{\textcolor{Maroon}{\small \ $\downarrow${#1}}}

\usepackage{pifont}% http://ctan.org/pkg/pifont
\newcommand{\cmark}{\textcolor{OliveGreen}{\text{\ding{51}}}}
\newcommand{\xmark}{\textcolor{Maroon}{\text{\ding{55}}}}

\title{\raisebox{-0.17em}{\includegraphics[height=1em]{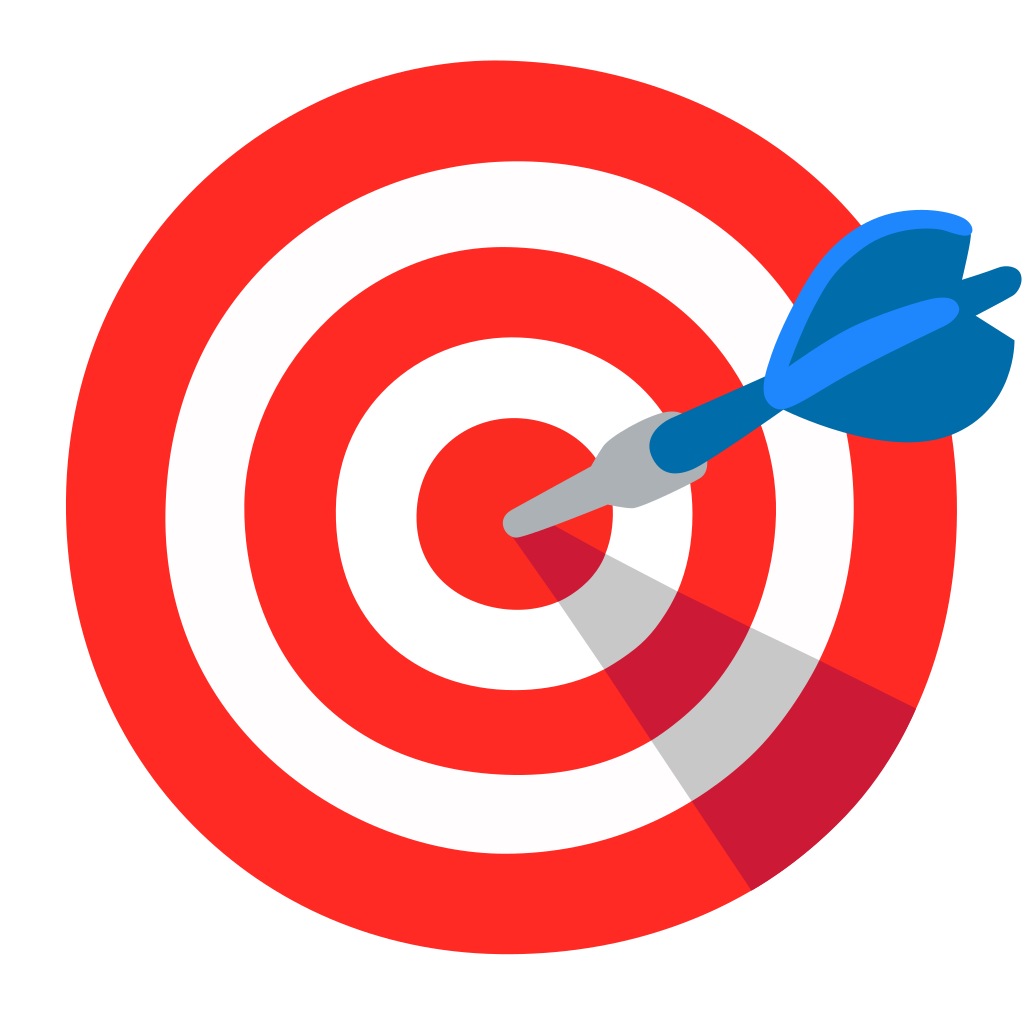}}
  \rawmethod[-Math]: \fullmethod\\
for Mathematical Problem-Solving}

\author{%
  Yuxuan Tong\thanks{Work done during visit to
  HKUST.}\hspace{4pt}$^1$, Xiwen Zhang$^{2}$, Rui Wang$^{2}$, Ruidong
  Wu$^{2}$, Junxian He$^{3}$ \\[5pt]
  $^1$Tsinghua University \quad $^2$Helixon Research \quad $^3$HKUST
  \\[3pt]
  \texttt{tongyx21@mails.tsinghua.edu.cn ~~ junxianh@cse.ust.hk} \\
}

\begin{document}

\maketitle

\input{abs}

\begin{figure*}[t!]
  \centering
  \begin{subfigure}[t]{0.49\textwidth}
    \centering
    \includegraphics[width=0.95\linewidth]{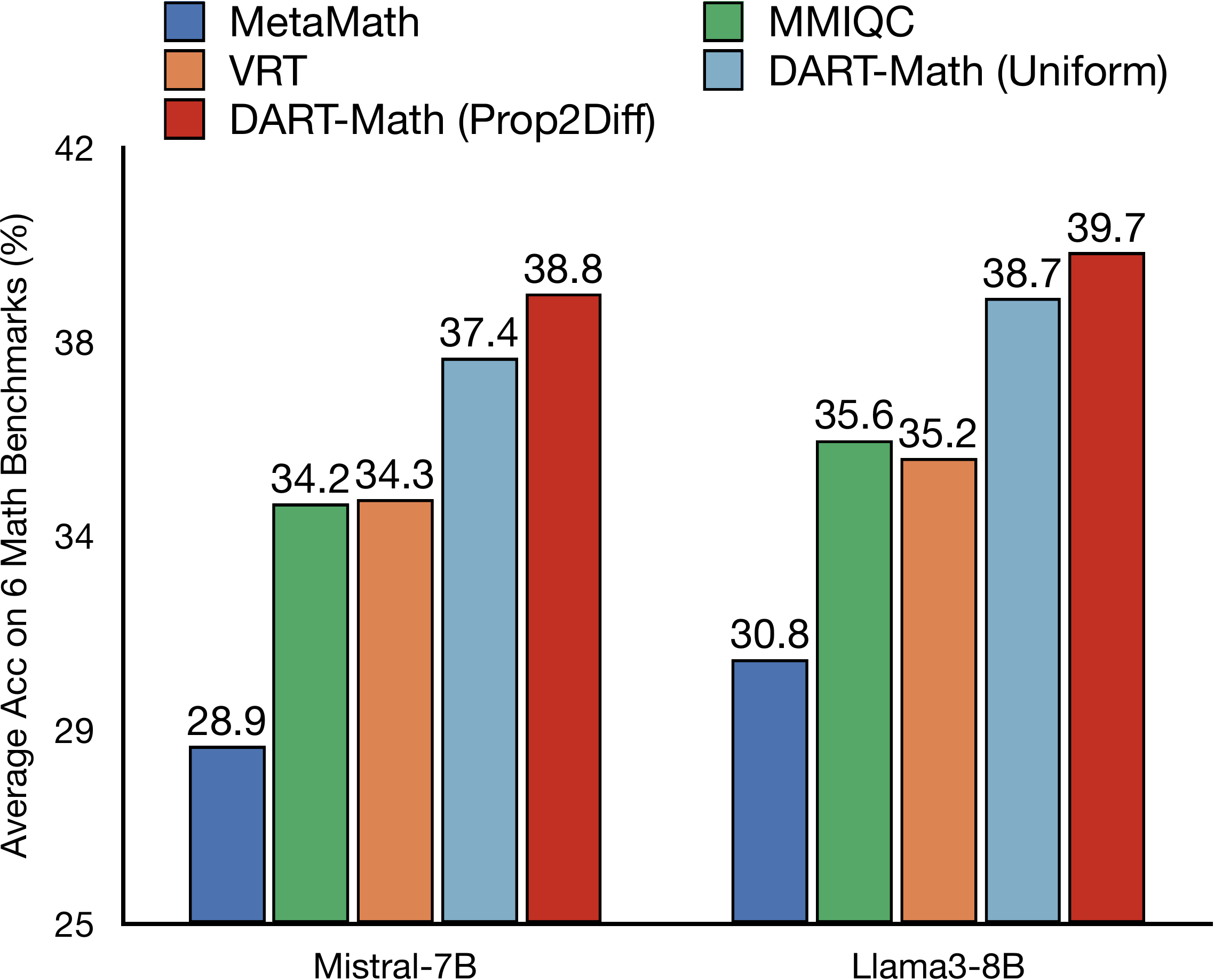}
  \end{subfigure}
  \hfill
  \begin{subfigure}[t]{0.48\textwidth}
    \centering
    \includegraphics[width=0.8\linewidth]{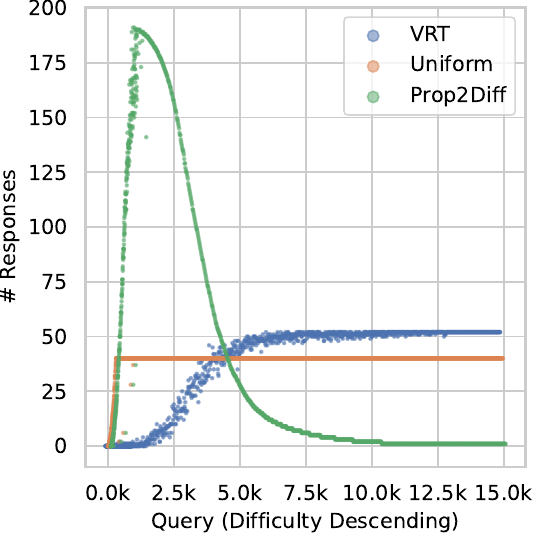}
  \end{subfigure}
  \caption{\textbf{Left:} Average accuracy on six mathematical
    benchmarks. We compare with models fine-tuned on the best, public
    instruction tuning datasets for mathematical problem-solving:
    MetaMath~\citep{yu2024metamath} with 395k examples,
    MMIQC~\citep{liu2024mmiqc} with 2.3 million examples, as well as
    vanilla rejection tuning (VRT) with 590k examples. Both DART-Math
    (Uniform) and DART-Math (Prop2Diff) use 590k training examples.
    \textbf{Right:} Number of responses for each query descending by
    difficulty across 3 synthesis strategies. Queries are from the MATH
    training split~\citep{hendrycks2021math}. VRT is the baseline
    biased towards easy queries, while Uniform and Prop2Diff are
    proposed in this work to balance and bias towards difficult queries
  respectively. Points are slightly shifted and downsampled for clarity.}
  \label{fig:main_perf_and_ds}
  \vspace{-15pt}
\end{figure*}

\input{intro}

\input{method}
\input{exp}

\input{discussion}
\section*{Acknowledgments}

We thank Zhiyuan Zeng, Wei Xiong and
Chenyang Zhao
for helpful discussions. Yuxuan is partially supported by
Tsinghua University
Initiative Scientific Research Program (Student Academic Research
Advancement Program).

\bibliography{ref}
\bibliographystyle{iclr2022_conference}

\clearpage

%%%%%%%%%%%%%%%%%%%%%%%%%%%%%%%%%%%%%%%%%%%%%%%%%%%%%%%%%%%%

\input{appendix}

%%%%%%%%%%%%%%%%%%%%%%%%%%%%%%%%%%%%%%%%%%%%%%%%%%%%%%%%%%%%

\input{checklist}

\end{document}

%% file: math_commands.tex
\usepackage{amsmath,amsfonts,bm}

\def\eqref#1{equation~\ref{#1}}
\def\1{\bm{1}}

\DeclareMathAlphabet{\mathsfit}{\encodingdefault}{\sfdefault}{m}{sl}
\SetMathAlphabet{\mathsfit}{bold}{\encodingdefault}{\sfdefault}{bx}{n}

%% file: abs.tex
\begin{abstract}

  Solving mathematical problems requires advanced reasoning abilities
  and presents notable challenges for large language models. Previous
  works usually synthesize data from proprietary models to augment
  existing datasets, followed by instruction tuning to achieve
  top-tier results. However, our analysis of these datasets reveals
  severe biases towards easy queries, with frequent failures to
  generate any correct response for the most challenging queries.
  Hypothesizing that difficult queries are crucial to learning complex
  reasoning, we propose \textit{\fullmethod} (\method), a method that
  allocates difficult queries more trials during the synthesis phase,
  enabling more extensive training on difficult samples.
  Utilizing \method, we have created new datasets for mathematical
  problem-solving that focus more on difficult queries and are
  substantially smaller than previous ones. Remarkably, our synthesis
  process solely relies on a 7B-sized open-weight model, without
  reliance on the commonly used proprietary GPT-4.
  We fine-tune various base models on our datasets ranging from 7B to
  70B in size, resulting in a series of strong models called \model.
  In comprehensive in-domain and out-of-domain evaluation on 6
  mathematical benchmarks, \model~outperforms vanilla rejection
  tuning significantly, being superior or comparable to previous
  arts, despite using much smaller datasets and no proprietary
  models. Furthermore, our results position our synthetic datasets as
  the most effective and cost-efficient publicly available resources
  for advancing mathematical problem-solving.\footnote[1]{Our datasets,
    models and code are publicly available at
  \url{https://github.com/hkust-nlp/dart-math}.}
  % , which are either superior to or competitive with previous arts,
  % despite using much smaller datasets and no proprietary models.
  % Furthermore, \model-DeepSeekMath-7B boosts MATH accuracy to
  % 53.6\%, outperforming reinforcement learning variants.
  % On various out-of-domain benchmarks, DART-Math models also
  % achieve superior or competitive performance.

\end{abstract}

%% file: intro.tex
\section{Introduction}

% General goal and limitation: Reasoning
% Reasoning is one of the fundamental abilities of human cognition to
% solve complex tasks, thus also essential for building AGI.
% Recent years have witnessed significant progress on various NLP
% tasks brought by LLMs. However, current LLMs are still limited in
% complex reasoning tasks in domains such as mathematics, programming
% and physics.
Recent years have seen remarkable advancements in various tasks
through the use of large language models
(LLMs)~\citep{brown2020language,
touvron2023llama,chowdhery2023palm,anthropic2023claude,openai2023gpt4}.
However, these models still struggle with complex
reasoning~\citep{hendrycks2021math,jimenez2024swebench,he2024olympiadbench,lin2024criticbench},
a cornerstone of human cognitive essential for tackling intricate
tasks. Mathematical reasoning, in particular, represents a
significant challenge and stands as one of the most difficult
categories of reasoning for state-of-the-art
LLMs~\citep{hendrycks2021math,cobbe2021training,zheng2022minif2f}.

In this work, we focus on mathematical problem-solving to explore
enhancement of the mathematical reasoning abilities of pretrained LLMs.
We investigate instruction tuning~\citep{flan, self-instruct}, which
is recognized as the most cost-effective method and achieves the
state-of-the-art performance on various mathematical
benchmarks~\citep{yu2024metamath,yue2024mammoth}.
% Prior research has improved the mathematical problem-solving
% abilities of pretrained LMs through
% prompting~\citep{wei2022chain,fu2022complexity}, search-based
% decoding~\citep{lightman2023verifystep,yao2024tree,xie2024self},
% and post-training techniques such as continual
% pretraining~\citep{azerbayev2023llemma,shao2024deepseekmath} and
% instruction tuning~\citep{yu2024metamath,yue2024mammoth}.
%  Among these, instruction tuning is recognized as the most
% cost-effective method, achieving the state-of-the-art performance
% on various benchmarks.
Current SOTA instruction tuning methods for mathematical
problem-solving are typically implemented as augmenting existing
training datasets with synthetic data generated from proprietary
models like GPT-4~\citep{openai2023gpt4}.
A prevalent method of data augmentation is to sample multiple
responses to given queries from a strong model and filter out the
incorrect ones.
% ~\citep{yuan2023rft,yu2024metamath,toshniwal2024openmathinstruct}.
% by comparing the synthetic answer to the ground-truth
% answers~\citep{yuan2023rft,yu2024metamath,toshniwal2024openmathinstruct}
% or other heuristic ways like sub-problem result
% consistency~\citep{huang2024kpmath}.
This method, known as rejection tuning, ensures the high quality of
the augmented thought steps and yields competitive
performance~\citep{yuan2023rft,yu2024metamath,singh2023restem}.

\begin{figure*}[t!]
  \centering
  \begin{subfigure}[t]{0.32\textwidth}
    \centering
    \includegraphics[width=\linewidth]{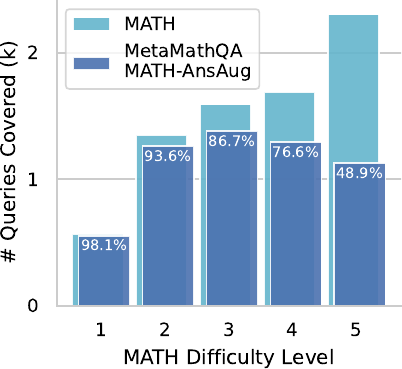}
  \end{subfigure}
  \hfill
  \begin{subfigure}[t]{0.32\textwidth}
    \centering
    \includegraphics[width=\linewidth]{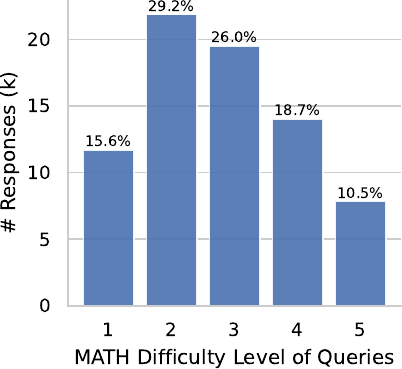}
  \end{subfigure}
  \hfill
  \begin{subfigure}[t]{0.32\textwidth}
    \centering
    \includegraphics[width=\linewidth]{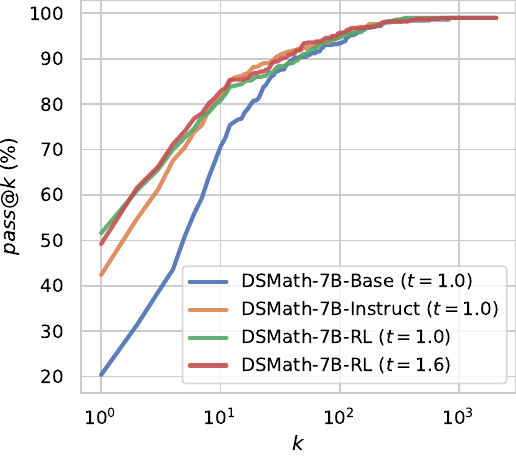}
  \end{subfigure}
  \caption{\textbf{Left:} Number of queries in the MATH training set
    and the MetaMathQA-MATH-AnsAug set across 5 difficulty levels
    annotated by humans. MetaMathQA-MATH-AnsAug is generated through
    rejection sampling from the original training queries. We annotate
    the query coverage ratio of MetaMathQA. While the most difficult
    queries (Level 5) are predominant in the original set, synthetic
    examples bias towards easier queries, dropping over 50\% of the
    most difficult queries. \textbf{Middle:} Total number of responses
    for queries across different difficulty levels in
    MetaMathQA-MATH-AnsAug. The most difficult queries represent the
    smallest proportion, only accounting for 10.5\% of all the samples.
    \textbf{Right:} $pass@k$ accuracy of different DeepSeekMath
    (DSMath) models and temperatures ($t$) on
    MATH500~\citep{lightman2023verifystep}, a subset of MATH test set.
    With enough trials, models are actually able to sample out
  answer-correct responses to most (>99\%) queries.}
  \label{fig:bias_pass_at_k}
  \vspace{-10pt}
\end{figure*}

%  The resulting instruction tuning is termed as \textit{rejection fine-tuning}.
However, after careful examination of these SOTA synthetic datasets,
we find that they suffer from a severe bias towards responses to easy
queries and low coverage for hard queries. For example, as shown in
Figure \ref{fig:bias_pass_at_k} (Left and Middle), while the original
queries vary in difficulty, the augmented samples in the MetaMathQA
dataset~\citep{yu2024metamath} focus more on easier queries, with
zero new responses generated for 51.1\% of the most difficult
training queries in the MATH training set~\citep{hendrycks2021math}.
This phenomenon commonly exists in rejection-sampling-based data
synthesis which typically samples \textit{an equal number of raw
responses for each query}, disadvantaging difficult queries that are
less likely to yield correct responses.
We hypothesize that such biases hinder the learning of mathematical
problem-solving, since difficult examples are often deemed more
crucial during training~\citep{sorscher2022beyond,burns2023weak,liu2024deita}.

To address this issue,
% we first examine whether LLMs are able to generate correct
% responses by drawing more samples, and find that the $pass@k$
% accuracy of xx model on the MATH benchmark improves from xx to xx
% when we increase $k$ from 1 to xxx, as shown in
% \autoref{fig:pass_at_k}\draft{https://www.notion.so/hkust-nlp/Paper-Draft-6d63bb91e5cf4e1b94fa04daf5da77c8?pvs=4\#b842e503190d454cb89a5f12791d24b9}.\jh{in
% the caption of this figure, make sure to cite xwin as well to
% mention it is concurrent finding}
% noticing that LLMs are able to generate correct response to hard
% queries with more trials, as shown in \autoref{fig:bias_pass_at_k} (Right),
we propose \textit{Difficulty-Aware Rejecting Tuning} (\method), a
method that prioritizes more sampling trials for challenging queries,
thereby generating synthetic datasets enriched with more responses
for difficult questions compared to previous methods.
Specifically, we develop two strategies to achieve this:
\textit{Uniform} which collects the same number of correct responses
for all queries, and \textit{Prop2Diff} which biases the data samples
towards the difficult queries, contrasting with vanilla rejection tuning.
These different strategies are summarized in
Figure~\ref{fig:main_perf_and_ds} (Right),
where the difficulty of a query is automatically assessed by sampling
multiple responses and calculating the ratio of incorrect answers.
Our difficulty-aware synthesis produces two synthetic datasets
corresponding to Uniform and Prop2Diff strategies respectively,
consisting of $\sim$590k examples.
% The resulting synthetic dataset contains 585k examples, with the
% number of responses for each query approximately proportional to
% its difficulty.
Notably, while previous works mostly utilize GPT-4 to synthesize
data, we only rely on the DeepSeekMath-7B-RL
model~\citep{shao2024deepseekmath} to produce all the data, thereby
eliminating dependence on proprietary models.

In our experiments, we evaluate \method[] based on
Mistral-7B~\citep{jiang2023mistral},
DeepSeekMath-7B~\citep{shao2024deepseekmath}, Llama3-8B, and
Llama3-70B~\citep{meta2023llama3}, creating a series of strong
mathematical models that termed \model.
Across 6 in-domain
% ~\citep{hendrycks2021math,cobbe2021gsm8k}
and challenging out-of-domain benchmarks,
% ~\citep{saxton2018dmmath,patel2021svamp,chen2023theoremqa,tang2024mathscale,he2024olympiadbench},
\model~significantly outperforms vanilla rejection tuning and the
baselines trained on the previously established top public datasets
as shown in Figure~\ref{fig:main_perf_and_ds} (Left), this is often
achieved with smaller training data size.
% achieves superior or comparable results to the state-of-the-art
% models that use significantly larger datasets and rely on GPT-4.
% Based on the DeepSeekMath-7B model, \method[] achieves 53.8\% on
% MATH~\citep{hendrycks2021math}, even outperforming the
% DeepSeekMath-7B-RL (53.2\%) variant and GPT-4-0314 (52.6\%).
For example,
\dataset~improves Llama3-8B from 21.2\% to 46.6\% on
MATH~\citep{hendrycks2021math}, and from 51.0\% to 82.5\% on
GSM8K~\citep{cobbe2021gsm8k};
% based on the DeepSeekMath-7B model, \model~achieves 53.8\% on MATH,
% even outperforming the DeepSeekMath-7B-RL (53.2\%) variant and
% GPT-4-0314 (52.6\%).
Our results mark the \dataset~datasets as the state-of-the-art
\textit{public} resources of instruction tuning for mathematical
problem-solving.

%% file: method.tex
\section{Biases in Rejection-Based Data Synthesis}
In this section, we first introduce the background for rejection
sampling and rejection tuning, and then present our examination on
the biases of rejection-based data synthesis.
% Then we propose our approach, a difficulty-aware method to
% synthesize data and fine-tune the models.
\subsection{Background: Rejection Sampling and Rejection Tuning}
\label{sec:problem}
% Here, we formalize a question about training data construction as following:

We begin by formulating the data synthesis setting used for
instruction tuning. For instruction tuning, the training dataset
consists of $(x,y)$ pairs, where $x$ is the input query and $y$ is the response.
The process of data synthesis involves generating new $(x,y)$ pairs
to augment the original training dataset, thereby enhancing performance.
For each input query $x_i$, it is typical to sample $M$ responses
from advanced models such as GPT-4, forming the set $\{(x_i,
y_i^{(j)})\}_{j=1}^M$.
In the context of mathematical problem-solving, a subsequent
filtering step is often implemented to eliminate incorrect
$y_i^{(j)}$. This elimination is based on whether the final answer in
the synthetic response aligns with the ground-truth
answer.\footnote{Strictly speaking, final answer correctness does not
  necessarily imply intermediate reasoning correctness. We do not make
further distinction across this paper which is not our focus.} % ,
% which is commonly observed in LLMs' output~\citep{lightman2023verifystep}
This is crucial as mathematical reasoning poses a significant
challenge for current LLMs, and the generated $y_i^{(j)}$ may often
be of poor quality.
This method of response sampling is known as \textit{rejection
sampling}, and the subsequent fine-tuning process is referred to as
\textit{rejection tuning}, which is widely employed to enhance the
mathematical problem-solving abilities of
LLMs~\citep{zelikman2022star,
yuan2023rft,yu2024metamath,singh2023restem,xu2024chatglmmath}.
In addition to response synthesis, the queries are typically kept
constant~\citep{singh2023restem,hosseini2024vstar,toshniwal2024openmathinstruct}
or altered in a controlled manner~\citep{yu2024metamath} to ensure
that ground-truth answers are readily available, which facilitates
the implementation of rejection sampling.
While some studies also synthesize queries without utilizing
rejection tuning~\citep{li2024common,tang2024mathscale}, our focus in
this work is primarily on rejection tuning, a method prevalently used
for advancing the mathematical skills of LLMs.
%  Suppose we are synthesizing the responses given a fixed query set
% $\{x_i\}_{i=1}^n$,
% In synthetic data regime, given a fixed query set $\sX$, for each
% query $x_{i} \in \sX$, let $y_{i}$ be the weight of its
% (answer-correct) responses in the final training set, $\vx_{i}$ be
% its metrics, what is the $f(\vx_{i})$ satisfying $y_{i} \propto
% f(\vx_{i})$ that produce the best performance? Here, we focus on
% query difficulty, which is estimated as query's pass rate by some
% agent $\pi$, i.e., $\vx_{i}=[x_{i,\pi}]$.

% This is meaningful because \texttt{VRT} can be succinctly
% formalized as \sample, where $k_{\text{max}}$ is the number of raw
% responses sampled on each query, which is statistically equivalent
% to \passrate because $P(Y)=\operatorname{Binomial}(k_{\text{max}},
% x_{i,\pi}) \to E(Y) = x_{i,\pi}$.
\subsection{On the Imbalance of Rejection-Based Data Synthesis}
\label{sec:imbalance}
Next, we examine a representative synthetic dataset to identify the
inherent biases present in rejection-based data synthesis as
implemented in most existing works.
Specifically, our analysis focuses on the AnsAug subset of the
MetaMathQA-MATH dataset~\citep{yu2024metamath},
which is a synthetic dataset that produces multiple responses for
each query in the original training set of the MATH
dataset~\citep{hendrycks2021math}, through rejection sampling as
described in \textsection\ref{sec:problem}.
MetaMathQA has been recognized as one of the most effective synthetic
datasets for mathematical problem-solving.
% yielding competitive performance.
We concentrate on the MATH split because it is a notably challenging
benchmark in mathematical reasoning, equipped with human-annotated
difficulty levels that aid in our analysis.
% Below, we delve into the response distributions in this synthetic dataset.
%  for different queries.

\paragraph{Rejection-based data synthesis biases towards easy queries:}
Across different difficulty levels, Figure~\ref{fig:bias_pass_at_k}
(Left) shows the original query distribution of the MATH training
dataset as well as the new query distribution after synthesis in the
MetaMathQA-Math dataset.
While the most difficult queries (Level 5) takes the largest
proportion in the original query set, MetaMathQA changes the query
distribution implicitly towards easier queries, dropping many hard problems.
For instance, the proportion of Level 5 (the most difficult) queries
notably decreases by 51.1\%, indicating that rejection sampling fails
to generate any correct response for those queries.
As a result, as depicted in Figure~\ref{fig:bias_pass_at_k} (Middle),
the responses to the most difficult queries only account for 10.5\%
of all the samples.
% in MetaMathQA-Math, fails to cover 51.1\% of the most difficult
% (level 5) queries, resulting in zero synthetic responses for xx\%
% of all the queries in total.
Such a phenomenon generally exists in datasets synthesized through
the conventional rejection sampling method outlined in
\textsection\ref{sec:problem}, primarily because \textit{the same
number of responses} is sampled for each query, yet the likelihood of
obtaining correct responses for difficult queries is significantly
lower, sometimes even zero.
We hypothesize that this bias towards easy queries could
substantially undermine the effectiveness of instruction tuning, as
hard queries are often considered critical for instruction
tuning~\citep{lu2024instag,liu2024deita}.
We note that this bias towards easy queries is less pronounced on
relatively simple datasets such as GSM8K~\citep{cobbe2021gsm8k},
where most queries are easier
% for example, GPT-4 can achieve over 90\% accuracy on GSM8K
and it is not difficult to sample correct responses for most of the queries.
However, the bias remains a significant concern when tackling
challenging tasks, which represent a more compelling and complex
field of study for LLMs.
Building on these findings, we will next introduce our method as a
potential remedy to the limitations of vanilla rejection tuning.

\section{\method[] --- \fullmethod}\label{sec:method}
% In this section, we present a difficulty-aware rejection sampling
% approach to synthesize data, mitigating the biases in vanilla
% rejection sampling.
% \jh{mention we focus on text-only synthesis without using tools}

\subsection{Open-Weight Models Are Able to Generate Good Responses}
Intuitively, we aim to collect a sufficient number of responses for
the difficult queries.
To assess whether this goal is achievable, given that models might
not generate correct responses for challenging queries despite
extensive sampling, we explore the capabilities of
DeepSeekMath-7B-RL~\citep{shao2024deepseekmath}, a strong model
specifically trained for mathematical reasoning.
Figure~\ref{fig:bias_pass_at_k} (Right) demonstrates the $pass@k$
accuracy on the queries in MATH500~\citep{lightman2023verifystep}, a
subset of MATH test set,
% similar in both difficulty level and domain distributions,
indicating the proportion of queries that have at least one correct
response when sampling $k$ responses for each query.
Notably, even though the synthesis model possesses only 7B
parameters, a 90\% $pass@k$ accuracy can be achieved when sampling
over 100 responses per query.
These results are consistent with the findings from recent
studies~\citep{toshniwal2024openmathinstruct,shao2024deepseekmath,li2024common},
which suggest that strong open-weight models are able to synthesize
correct responses for most of the queries.
This evidence supports the potential for effectively mitigating the
insufficient coverage for difficult queries through strategic
response sampling, which we introduce next.
% Specifically, we adopt
% Inspired
% by~\citet{toshniwal2024openmathinstruct,shao2024deepseekmath} which
% find that open-weight models such as
% Mixtral-8x7B~\citep{jiang2024mixtral} and
% DeepSeekMath-7B-RL~\citep{shao2024deepseekmath} are able to
% generate correct responses when sampling larger number of response,
% achieving high $pass@k$ accuracies that denote the ratio of queries
% that have at least one correct response.

\subsection{\sampling[] --- Difficulty-Aware Rejection Sampling}
\label{sec:dars}
Motivated by the observation above, we aim to collect more responses
for harder queries. Specifically, we introduce two strategies to
increase the number of correct responses for difficult queries: (1)
\textbf{Uniform}, which involves sampling responses for each query
until each query accumulates $k_u$ correct responses, and $k_u$ is a
preset hyperparameter determined by the desired size of the synthetic
dataset; (2) \textbf{Prop2Diff}, where we continue sampling responses
until the number of correct responses for each query is (linearly)
proportional to its difficulty score. The most challenging queries
will receive $k_p$ responses and $k_p$ is a hyperparameter.
This method introduces a deliberate bias in the opposite direction to
vanilla rejection sampling, towards more difficult queries.
Prop2Diff is inspired by previous works that demonstrate difficult
queries can be more effective to enhance model
capabilities~\citep{sorscher2022beyond,liu2024deita}.
Both the Uniform and Prop2Diff strategies prescribe a specific number
of correct response for each query, determined by $k_u$ or $k_p$.
Nevertheless, there are certain queries which we cannot sample out
the designated number of correct responses even with extensive
sampling efforts. To avoid endless running of the synthesis, we
impose a cap on the maximum allowable number of raw samples per query
as $n_{\max}$ --- once this limit is reached for a particular query,
we cease further sampling and retain any correct responses that have
been gathered.
The straightforward implementation of the Prop2Diff strategy risks
generating no synthetic responses for easier queries if $k_p$ is set
small. To mitigate this, we guarantee at least one synthetic response
for each query when implementing Prop2Diff.
While it might seem sufficient to rely on the original, real training
dataset to ensure at least one human-annotated response per query,
our findings highlight the importance of maintaining synthetic
response coverage to learn to solve easy problems, as we will
quantitatively shown in \textsection\ref{sec:analysis}, partially
because the human-annotated response is less detailed and not as
beneficial as synthetic responses, demonstrated previously
in~\citet{yu2024metamath}.
For both Uniform and Prop2Diff strategies, we use the
DeepSeekMath-7B-RL model to synthesize responses.
We refer to the two sampling strategies as \darsu~and \darsp~respectively.
% We refer to the datasets constructed via these methods as
% \dataset[-Uniform] and \dataset[-Hard] respectively.
Though most previous methods are difficulty-agnostic, a few methods
try assigning more budget to more complex questions to boost coverage,
such as ToRA \citep{gou2024tora} and MARIO \citep{liao2024mario}. However,
ToRA/MARIO mainly focus on improving coverage without managing the
distribution explicitly, leading to datasets that may still bias
towards easy queries, while \sampling[] explicitly controls the final
distribution of the training dataset, completely eliminating the bias
and also achieving higher coverage on the hardest queries. For more
details about the comparison, we refer readers to Appendix~\ref{app:comparison}.
As \darsp~requires assessing difficulties of queries, next we introduce
an automatic approach to measure difficulties.

\paragraph{Evaluating Difficulty:}
Previous studies have used proprietary models like ChatGPT to assess
the difficulty or complexity of data samples~\citep{lu2024instag,liu2024deita}.
In this work, we introduce a new metric, \textit{fail rate} --- the
proportion of incorrect responses when sampling $n_{d}$ responses for
a given query --- as a proxy for difficulty.
This metric aligns with the intuition that harder queries less
frequently yield correct responses.
% Fail rate corresponds to the intuition that harder queries are
% rarer to encounter correct responses.
We utilize DeepSeekMath-7B-RL as the sampling model to evaluate
difficulty across all experiments in the paper. Varying this sampling
model to align with the generative model may further enhance
performance, which we leave as future work.
Notably, one of the benefits of fail rate is that it allows to reuse
the sampled responses during difficulty evaluation as synthetic
responses for dataset construction. See implementation details in
Appendix~\ref{app:synthesis}.

\input{tables/datasets}
\subsection{The \dataset[] Datasets}
\label{sec:dart-data}
% \jh{to revise}\jh{mention that GSM8K is simple, do not have obvious
% bias and dart plays a smaller role there}
We utilize \darsu~and \darsp~to construct two datasets,
\dataset[-Uniform] and \dataset[-Hard] respectively for instruction tuning.
We use the original training queries of the
GSM8K~\citep{cobbe2021gsm8k} and MATH datasets to synthesize responses.
We maintain fixed queries to better isolate the effects of
difficulty-aware rejection tuning, while techniques for query
augmentation, as discussed in prior studies~\citep{yu2024metamath},
could be potentially incorporated to further improve the performance.
The synthetic datasets are augmented with the original GSM8K and MATH
training data to form the final datasets.
We set $k_u$ in \darsu~as 40 and $k_p$ in \darsp~as 192 to form both
datasets of around 590k samples.
Our data samples only involve natural language reasoning without
using external tools such as code execution.
Comparison of our datasets with previous datasets is illustrated in
Table~\ref{tab:datasets}.
Our datasets are generally smaller than most previous datasets, and
in \textsection\ref{sec:res} we will empirically demonstrate that
\textbf{the DART datasets are the most cost-effective datasets
publicly available}.
Remarkably, our approach solely utilizes DeepSeekMath-7B-RL to
evaluate difficulty of queries and synthesize responses, without
relying on ChatGPT that is commonly used in other studies.

Our approach typically requires more sampling trials than vanilla
rejection sampling to generate a dataset of comparable size because
difficult queries often need more samples to secure the required
number of correct responses. Despite this, it is crucial to point out
that our overall training cost does not exceed that of vanilla
instruction tuning.
% For convenience, when constructing datasets of different difficulty
% distributions, we implement as first use \texttt{DARS} to sample
% enough answer-correct responses to every query and then selecting
% subsets satisfying different $y_{i}=f(x_{i,\pi})$ from the pool.
% First, we try to sample 192 (limited by computation resources)
% answer-correct responses to every query in MATH training set. For
% the metric of dataset size, we use the maximum number of responses
% to one query $k_{\text{max}} \in \{2,4,8,12,16,24,32,64,128,192\}$
% (which is proportional to the most common metric $|\sQ|$), which
% allows succinct expression of number of responses to different
% queries $k_{i}$. Specifically, for each $y_{i}=f(x_{i,\pi})$ except
% for \sample, let $y_{\text{max}}$ be the maximum weight, for each
% query $q_{i} \in \sQ$, we calculate $k_{i}=k_{\text{max}} \cdot
% \nicefrac{y_{i}}{y_{\text{max}}}$, and randomly sample $k_i$
% samples from its answer-correct responses. Specifically, for the
% same $y_{i}=f(x_{i,\pi})$, we control that smaller datasets are
% subsets of bigger datasets. For \sample, for each $k_{\text{max}}$
% we simulate VRT by sampling the number of answer-correct responses
% from $\operatorname{Binomial}(k_{\text{max}}, x_{i,\pi})$, where
% $\pi=\text{DeepSeekMath-7B-RL}$ for each query $q_{i}$. Note that
% some queries are so difficult that even millions of raw responses
% do not contain enough number of answer-correct responses, which
% will cause the final datasets not to strictly conform
% $y_{i}=f(x_{i,\pi})$, but the trends of difficulty distribution
% still hold true.
We emphasize that the data synthesis process is a one-time effort.
Once the synthetic dataset is created, it can be utilized for
multiple training runs across various base models. Furthermore, this
dataset will be publicly available, extending its utility to a wide
range of users.
% interested in enhancing the mathematical problem-solving
% capabilities of their models.
From this perspective, the initial higher synthesis cost is
effectively amortized over numerous training runs and the broad user
base, rendering the synthesis cost virtually imperceptible to
individual dataset users.
We will discuss the synthesis cost further in \textsection\ref{sec:analysis}.

%% file: tables/datasets.tex
% Please add the following required packages to your document preamble:
% \usepackage{booktabs}
\begin{table}[t]
  \centering
  \resizebox{0.9\textwidth}{!}{
    \begin{tabular}{@{}lrlc@{}}
      \toprule
      Dataset            & \# Samples (k) & Synthesis Agent     &
      Open-Source \\ \midrule
      % ToRA-Corpus~\citep{gou2024tora} & 16 & GPT-4 & \xmark \\
      WizardMath~\citep{luo2023wizardmath} & 96 & GPT-4 & \xmark \\
      MetaMathQA~\citep{yu2024metamath}         & 395          &
      GPT-3.5             & \cmark      \\
      % MAmmoTH~\citep{yue2024mammoth} & 262 & GPT-4+Human & \cmark \\

      MMIQC~\citep{liu2024mmiqc}              & 2294         &
      GPT-4+GPT-3.5+Human & \cmark      \\
      Orca-Math~\citep{mitra2024orcamath} & 200 & GPT-4 & \cmark \\
      Xwin-Math-V1.1~\citep{li2024common}  & 1440         & GPT-4
      & \xmark      \\
      KPMath-Plus~\citep{huang2024kpmath}  & 1576         & GPT-4
      & \xmark      \\
      MathScaleQA~\citep{tang2024mathscale} & 2021 & GPT-3.5+Human & \xmark \\
      % Math-Plus~\citep{yue2024mammoth2} & 894 & GPT-4+GPT-3.5
      % & \xmark      \\ \midrule
      \midrule
      \dataset[-Uniform]  & 591          & DeepSeekMath-7B-RL  & \cmark      \\
      \dataset[-Hard]     & 585          & DeepSeekMath-7B-RL  &
      \cmark      \\ \bottomrule
  \end{tabular}}
  \caption{Comparison between our \dataset~datasets and previous
    mathematical instruction tuning datasets. Most of previous
    datasets are constructed with ChatGPT, and many of them are not
    open-source, especially for ones of the best performance.
    % Our datasets only utilize a 7B open-source model and are open-source.
  }
  \label{tab:datasets}
  \vspace{-15pt}
\end{table}

%% file: exp.tex
\section{Experiments}

\subsection{General Setup}
Below we summarize the key setup details, while we include more
information in Appendix~\ref{app:exp_setup}.
\paragraph{Data synthesis:}
We synthesize responses using the original training queries of the MATH
and GSM8K datasets.
As described in \textsection\ref{sec:dars}, we utilize
DeepSeekMath-7B-RL to synthesize all the data.
We use temperature sampling with adjusted temperature to sample
answer-correct responses to difficult queries.
% Note that temperature value is not transferable between different
% models. See \textsection\ref{sec:temperature} for more details
% about sampling temperature.
We set the maximum number of output tokens as 2048 and adopt top-p
sampling with $p=0.95$.
We use chain-of-thought prompt~\citep{wei2022chain} to synthesize.
% required by the DeepSeekMath
% team\footnote{\url{https://huggingface.co/deepseek-ai/deepseek-math-7b-rl}}.
We use the vLLM library~\citep{kwon2023vllm} to accelerate the
generation. In our setting, sampling 35k samples on MATH / GSM8k
queries takes about 1 NVIDIA A100 GPU hour.

\paragraph{Training:}
We perform standard instruction tuning on our synthetic datasets
\dataset[-Uniform] and \dataset[-Hard], based on several base models
including Llama3-8B~\citep{meta2023llama3},
Mistral-7B~\citep{jiang2023mistral}, and Llama3-70B as
representatives of general models, and
DeepSeekMath-7B~\citep{shao2024deepseekmath} as the representative of
math-specialized models.
% We fine-tune the base models above on all the datasets constructed.
% We follow the standard supervised fine-tuning (SFT) pipeline based
% on teacher forcing~\citep{toomarian1992learning}, masking out the
% loss of the prompt tokens~\citep{radford2019gpt2}.
For simplicity, we keep most hyperparameters the same across
different models and datasets, and tune only several key
hyperparameters like learning rate and number of epochs, as detailed
in Appendix~\ref{app:train}.
%  See Appendix~\ref{app:train} for details about the training
% hyper-parameters and implementation details.

\paragraph{Evaluation:}
For comprehensive assessment of mathematical reasoning of the models,
we adopt 6 benchmarks for both in-domain and out-of-domain (OOD)
evaluation. Specifically, we use the GSM8K and MATH test set as the
in-domain test. GSM8K consists of  grade school arithmetic tasks and
are considered much simper than MATH that contains challenging
competition mathematical problems. For OOD test, we utilize the
following four challenging benchmarks:
% \begin{enumerate}
%     \item \textbf{CollegeMath}~\citep{tang2024mathscale}
% \end{enumerate}
\begin{itemize}
    \vspace{-5pt}
  \item \textbf{CollegeMath}~\citep{tang2024mathscale}: This test set
    contains 2818 college-level mathematical problems extracted from
    9 textbooks across 7 domains such as linear algebra and
    differential equations, testing generalization on complex
    mathematical reasoning in diverse domains.
  \item \textbf{DeepMind-Mathematics}~\citep{saxton2018dmmath}: This
    test set contains 1000 problems
    % generated with
    % code\footnote{\url{https://github.com/google-deepmind/mathematics_dataset}}
    from a diverse range of problem types based on a national school
    mathematics curriculum (up to age 16), testing basic mathematical
    reasoning in diverse domains.
  \item \textbf{OlympiadBench-Math}~\citep{he2024olympiadbench}: This
    benchmark contains 675 Olympiad-level mathematical problems from
    competitions, which is a text-only English subset of
    OlympiadBench, testing generalization on the most complex
    mathematical reasoning.
  \item \textbf{TheoremQA}~\citep{chen2023theoremqa}: This benchmark
    contains 800 problems focused on utilizing mathematical theorems
    to solve challenging problems in fields such as math, physics and
    engineering, testing generalization on theoretical reasoning in
    general STEM.
    % \item \textbf{SVAMP}~\citep{patel2021svamp}: This benchmark
    % contains 1000 arithmetic word problems with grade level up to 4
    % by applying simple variations over existing word problems. This
    % benchmark is similar to GSM8K and relatively easy compared to others.
    % aimed for testing generalization on basic arithmetic reasoning.
\end{itemize}

All results are from natural language reasoning without using
external tools, through greedy decoding.

\paragraph{Baselines:}
We compare \method[] with the state-of-the-art instruction-tuned
mathematical models such as MetaMath~\citep{yu2024metamath},
MMIQC~\citep{liu2024mmiqc}, KPMah-Plus~\citep{huang2024kpmath}, and
Xwin-Math~\citep{li2024common}.
We copy the results directly from the respective papers except for
MetaMath and MMIQC, where we run our own training since their
datasets are public.
As shown in Table~\ref{tab:datasets}, these SOTA datasets all rely on
proprietary models for data synthesis.
Another ablation baseline to \method[] is vanilla rejection tuning
(VRT), where we synthesize a dataset of the same size of 0.59M
examples with DeepSeekMath-7B-RL, using vanilla rejection sampling as
described in \textsection\ref{sec:problem}.
% VRT serves as an ablation baseline to our approach.
We note that there are other strong models such
as~\citet{yue2024mammoth,gou2024tora} that are trained to solve
mathematical problems utilizing code execution, we exclude them since
this study focuses on reasoning without using tools.

% \subsubsection{Evaluation Generation Configuration}

\input{tables/main_results}
\subsection{Main Results}
\label{sec:res}
\paragraph{Comparing with Vanilla Rejection Tuning:}
The main results are in Table~\ref{tab:main-results}.
\model~based on all four different base models outperforms the VRT
baselines on most benchmarks consistently.
Focusing on performance with 7-8B general base models,
\model[-Llama3-8B] (Uniform) surpasses the VRT baseline across all 6
benchmarks by an average of 3.5 absolute points, while
\model[-Llama3-8B] (Prop2Diff) achieves an average improvement of 4.5 points.
On the in-domain challenging MATH benchmark, \model~(Prop2Diff)
enhances performance over VRT by nearly 7 absolute points for both
Mistral-7B and Llama3-8B models. For OOD benchmarks,
\model~(Prop2Diff) shows particularly notable gains on more difficult
benchmarks, with improvements ranging from 5.2 to 9.5 absolute points
on CollegeMath, DeepMind-Mathematics, and OlympiadBench-Math. This
indicates effective generalization of our approach.
% The observations are similar on Mistral-7B.
These improvements over the VRT baselines demonstrate the
effectiveness of the proposed difficulty-aware rejection sampling.
We note that \model~does not greatly boost the relatively simple,
in-domain GSM8K benchmark.
This is expected, as explained in \textsection\ref{sec:imbalance},
because vanilla rejection tuning expected
% already performs well with simpler queries and
does not face severe bias issues like those seen in more challenging
datasets. Thus, difficulty-aware rejection sampling has a limited
impact on easy datasets.
% When scaling to Llama3-70B, \jh{todo}
Interestingly, on much stronger base models DeepSeekMath-7B and
Llama3-70B, the improvement margin of \model~over VRT narrows, with
about a 1-point gain on average.
We hypothesize that this is due to these models' extensive
pretraining on mathematical content.
This pretraining likely covers most skills that could be learned from
the GSM8K and MATH training queries, suggesting that
% our datasets based on these queries may not significantly enhance
% the models' capabilities beyond their current state, no matter how
% we synthesize responses.
% In such cases,
the query set itself, rather than the responses, becomes the
bottleneck. Thus augmenting the range of queries could be a more
effective strategy for future improvements.
%  and such pretraining already covers most of the abilities can be
% learned from GSM8K and MATH training queries, thus the GSM8K and
% MATH training queries --- which our datasets are based on --- may not
% add much new to models' abilities no matter how we synthesize
% responses for these queries.
% In this case, the query set, rather than the responses, becomes the
% bottleneck, and augmenting queries is potentially a more effective approach.
% Our analysis on scaling behaviors in \textsection\ref{sec:analysis}
% supports this hypothesis, showing that DeepSeekMath-7B can easily
% achieve over 50 accuracy on MATH with fewer than 10k samples, and
% two orders of magnitude increase of the data samples only raise the
% accuracy by 3 points, implying the limited benefits of further
% training on GSM8K and MATH queries.

\paragraph{Comparison with previous top-performing methods:}
\model~achieves superior or comparable performance to previous best models.
Specifically, when compared with MetaMath, \model~wins greatly in all cases.
Additionally, \model[-DSMath-7B] achieves the state-of-the-art
results for models sized 7-8B on challenging benchmarks such as MATH,
OlympiadBench-Math, and TheoremQA.
On average, \model[-Mistral-7B] (Prop2Diff) surpasses
Mistral-7B-MMIQC by 4.6 absolute points, despite using only a quarter
of its training sample size.
Compared with concurrent work KPMath-Plus which relies on GPT-4 and
has not released either the data or the model, our approach slightly
underperforms on Mistral-7B for GSM8K and MATH. However,
\model~excels against it on DeepSeekMath-7B by a significant margin,
utilizing around one-third of its training data size.
The Xwin-Math models perform well on the GSM8K benchmark but fall
behind \model~(Prop2Diff) on other challenging benchmarks overall,
particularly with a more pronounced gap on 70B models --- although we
note that their models are based on Llama2 which is not very fair to
compare with.
% We note that the comparison with
Importantly, we fully open-source our datasets and models,
designating both \dataset[-Uniform] and \dataset[-Hard] as the
\textbf{best-performing and most cost-effective public instruction
tuning datasets available for advancing mathematical problem-solving.}

\paragraph{Additional results:} For additional results, such as
domain-wise performance on MATH and
comparison to RL, we refer readers to Appendix~\ref{app:additional_results}.

\begin{figure*}[t!]
  \includegraphics[width=\linewidth]{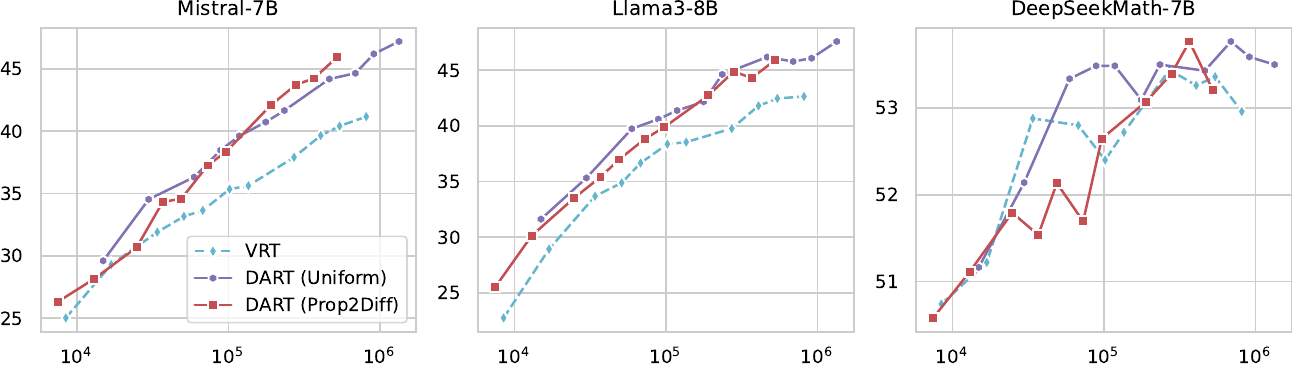}
  \centering
  \caption{
    Scaling curves of MATH test performance against number of
    training samples synthesized from MATH training queries, training
    is on three base models.
    %  See detailed explanations of legends in \textsection\ref{sec:problem}.
  }
  \label{fig:math_scaling}
  \vspace{-10pt}
\end{figure*}

\subsection{Analysis}
\label{sec:analysis}
\paragraph{Scaling behaviors of different data synthesis methods:}
We study the scaling behaviors of our data synthesis approach and
compare it to vanilla rejection sampling.
As described in~\ref{sec:imbalance}, our method is motivated to
mitigate the bias towards easy queries that are only pronounced in
challenging datasets.
%  but less obvious in simpler datasets such as GSM8K.
Therefore, in the scaling experiment we only synthesize responses for
the training queries of the challenging MATH dataset and report the
performance on the MATH test set.
Figure~\ref{fig:math_scaling} presents the results across three
different base models as we scale the training data size from
thousands to nearly 1 million samples.
We observe a steady improvement in performance as the training data
size increases exponentially. \method[] consistently outperforms VRT on
general base models Mistral-7B and Lllama3-8B, achieving better scaling.
On DeepSeekMath-7B, however, the performance differences between
various approaches are minimal. Observing the absolute accuracy
changes, DeepSeekMath-7B already achieves over 50\% accuracy with
just thousands of training samples, and scaling up to 1 million
samples leads to only a modest 3-point improvement. This is in stark
contrast to the over 20-point improvements seen on other models like
Mistral-7B and Llama3-8B.
As discussed in \textsection\ref{sec:res}, we believe this phenomenon
is due to the MATH training queries not being particularly beneficial
for DeepSeekMath-7B, which has undergone extensive math-specific
continual pretraining. Consequently, for DeepSeekMath-7B, the
differences between these approaches are not significant, and the
main bottleneck shifts to query coverage rather than the responses themselves.

\begin{figure*}[t!]
  \centering
  \begin{subfigure}[t]{0.48\textwidth}
    \centering
    \includegraphics[width=\linewidth]{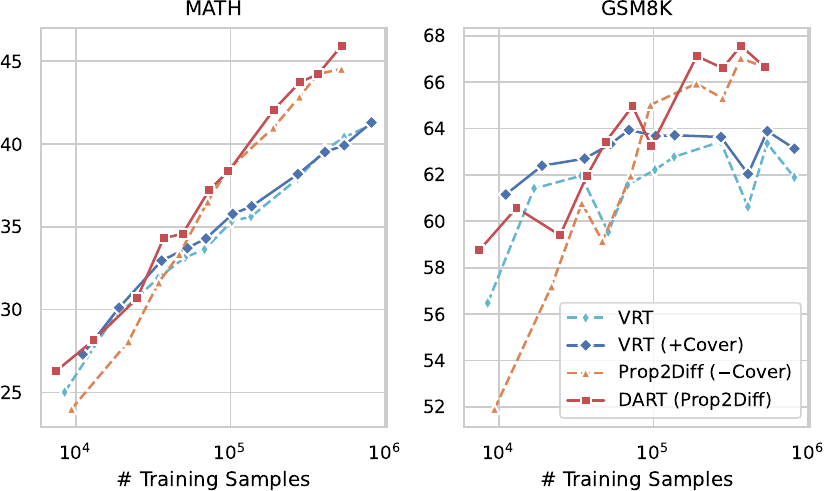}
  \end{subfigure}
  \hfill
  \begin{subfigure}[t]{0.48\textwidth}
    \centering
    \includegraphics[width=\linewidth]{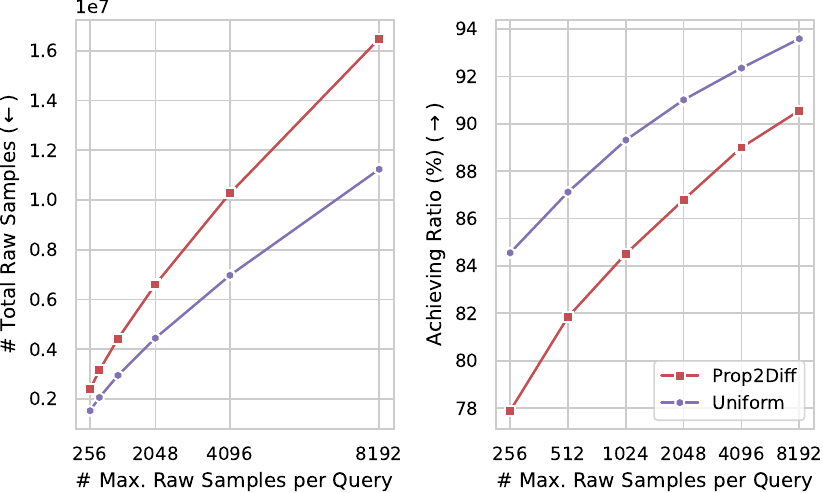}
  \end{subfigure}
  \caption{\textbf{From Left to Right, (1) and (2):}
    Scaling curves studying the effect of one-response coverage.
    ``Prop2Diff ($-$Cover)'' denotes \darsp~without enforcing at
    least one synthetic response for each query, while ``VRT
    (+Cover)'' denotes vanilla rejection sampling enforcing at least
    one synthetic response for each query.
    \textbf{(3) and (4):} The total number of raw samples needed, and
    the actual ratio ($r$) of queries achieving the desiderata of the
    two \sampling[] synthesis strategy for 585k-sized dataset curation
    respectively, when we vary the maximum allowable raw samples per
    query ($n_{\max}$).
    % Let $n_\textbf{max}$ be the maximum number of raw samples on each query,
    % Then total number of raw samples needed to curate a 585k-sized
    % dataset v.s. the maximum allowable raw samples per query
    % ($n_{\max}$). \textbf{(4)} The actual ratio of queries in the
    % synthetic dataset that meets the desiderata of the synthesis strategy.
    % exactly achieving $k_{i}$ correct responses is approximately
    % linear to $\log{n_{\max}}$,
    % thus total number of raw samples is approximately linear to the
    % successfully achieving ratio.
    % % a \texttt{Prop2Diff} / \texttt{Uniform} dataset is
    % approximately linear to $n_{\max}$,
    % while the actual ratio of queries exactly achieving $k_{i}$
    % correct responses is approximately linear to $\log{n_{\max}}$,
    % thus total number of raw samples is approximately linear to the
    % successfully achieving ratio.
  }
  \label{fig:cover_and_sample_cost}
  \vspace{-10pt}
\end{figure*}

\paragraph{Effect of one-response coverage:}
In \textsection\ref{sec:dars},
we describe that \darsp~can cause zero synthetic responses for easy
queries, especially when the number of training samples is small.
Therefore,
we ensure that the easy queries have at least one correct response
practically. Here we examine the impact of this one-response coverage
by comparing the Prop2Diff strategy with and without this coverage
constraint, as training data sizes increase.
Figure~\ref{fig:cover_and_sample_cost} (Left)
displays the outcomes on the MATH and GSM8K benchmarks respectively.
As anticipated, when the training data size is relatively small, the
one-response coverage proves beneficial, particularly on the simpler
GSM8K benchmark, improving accuracy by about 8 points. This suggests
that effective learning for easy problem-solving can be achieved with
just one additional correct response.
As we scale up the training data size, the natural increase in
coverage for easy queries causes that the difference between the two
approaches diminishes.
Additionally, we explore the implementation of one-response coverage
in vanilla rejection tuning to determine if adding one synthetic
response for difficult queries could address its issue of low
coverage for such queries. However, this modification does not
significantly aid in learning difficult queries, as observed on the
challenging MATH benchmark. This indicates that complex problems
generally require a greater number of training samples for effective learning.

\paragraph{Synthesis cost:}
\method[] generally needs more sampling trials to synthesize the same
size of dataset compared to vanilla rejection tuning, as discussed in
\textsection\ref{sec:dart-data}.
% since \method[] aims to collect more correct responses for the
% difficult queries.
It is important to underline that the synthesis cost, although
initially higher, is a one-time expense. Once the dataset is
synthesized, it can be used by the community and us to train numerous
models, effectively amortizing the cost.
% As shown in the main results of Table~\ref{tab:main-results}, the
% \dataset~dataset is the most cost-effective dataset publicly
% available to advance mathematical problem solving --- by
% synthesizing only once, other users in the community and us are
% able to train numerous models on the dataset, which actually
% amortizes the synthesis cost.
To provide a quantitative understanding of the synthesis cost, we
consider two main factors: $n_{\max}$, the maximum allowable raw
samples for each query, and $r$, the ratio of queries that achieve
the designated number of responses. If $n_{\max}$ is set too high,
sampling may continue indefinitely for particularly difficult or
noisy queries, resulting in a high synthesis cost. Conversely, a too
small $n_{\max}$ may result in many queries not gathering the
sufficient number of correct responses, leading to a lower $r$.
% --- essentially a small fraction of difficult queries are
% responsible for most of the synthesis cost in \method.
%
Figure~\ref{fig:cover_and_sample_cost} (Right) illustrates the total number
of raw samples required to synthesize 585k examples and the query
achieving ratio $r$ as we increase $n_{\max}$. When $n_{\max}$
reaches 2048, over 90\% of the queries can collect the designated
number of responses under \darsu, with a corresponding total number
of samples around 5 million.
To reach 90\% achieving ratio for \darsp, $n_{\max}$ needs to be at
least 8K, and the total number of raw samples exceeds 15 million.
In our experiments, we achieved an over 95\% ratio $r$, sampling
approximately 150 million samples in total, which required running
inference of DeepSeekMath-7B-RL for about 160 NVIDIA A100 GPU days.
Besides that synthesis is a one-time cost, we would like to emphasize
the number of samples is not a fair metric to compare synthesis cost
between different works --- our synthesis model of 7B size is
relatively inexpensive and fast to run, compared to the much more
costly and slower GPT-4 used in most previous studies.
Moreover, achieving a query ratio as high as 95\% may not be
necessary to reach good performance. A slightly lower ratio of 85\%
or 90\% might not significantly impact performance but could
substantially reduce the synthesis cost. We plan to explore this
balance further in future work.

% To validate the effectiveness of \method on a larger scale, we
% carry out additional experiments by extending the query set, base
% models and evaluation benchmarks. Specifically, we extend the query
% set to MATH and GSM8k training query set, add
% Llama-3-70B~\citep{meta2023llama3} as one of the largest and
% strongest open-source model to the base models and add the
% following out-of-domain evaluation benchmarks: the English
% competition subset of OlympiadBench (abbreviated as Olympiad-Math
% below)~\citep{he2024olympiadbench},
% DeepMind-Mathematics~\citep{saxton2018dmmath},
% TheoremQA~\citep{chen2023theoremqa} and
% SVAMP~\citep{patel2021svamp}. Limited by computation resources, we
% only train the models on the largest datasets of \failrate-Cover,
% \const, \sample.

% The results are shown in
% \autoref{tab:main-results}\draft{https://www.notion.so/hkust-nlp/Table-1-Main-results-6890ea5387354b78b783815813bab392?pvs=4\#6e1bc9dbf60849c3ad7180f9cfbb17f5}.

% \failrate-Cover and \const are significantly better than the
% \sample baseline, validating the importance of enough responses to
% hard queries.

% Accross different sizes and base model types, our \model achieve
% state-of-the-art on most mathematical reasoning benchmarks within SFT models.

%% file: tables/main_results.tex
\begin{table}[t]
  \resizebox{\textwidth}{!}{
    \begin{tabular}{lccccccccc}
      \toprule
      \multicolumn{1}{c}{\multirow{2}{*}{Model}}                &
      \multirow{2}{*}{\# Samples} & \multicolumn{2}{c}{In-Domain}
      & \multicolumn{4}{c}{Out-of-Domain}
      &              \\
      \multicolumn{1}{c}{}
      &       & MATH                  & GSM8K                 &
      College               & DM                    & Olympiad
      & Theorem               & AVG                   \\
      \midrule
      GPT-4-Turbo (24-04-09)
      & --    & 73.4                  & 94.5                  & --
      & --                    & --                    & 48.4
      & --                    \\
      GPT-4 (0314)
      & --    & 52.6                  & 94.7                  & 24.4
      & --                    & --                    & --
      & --                    \\
      Claude-3-Opus
      & --    & 60.1                  & 95.0                  & --
      & --                    & --                    & --
      & --                    \\
      Gemini 1.5 Pro
      & --    & 67.7                  & --                    & --
      & --                    & --                    & --
      & --                    \\
      \midrule
      \multicolumn{10}{c}{\textbf{70B General Base Model}} \\
      Llama2-70B-Xwin-Math-V1.1$^\dagger$
      & 1.4M  & 52.5                  & 90.2                  & 33.1
      & 58.0                  & 16.3                  & 14.9
      & 44.2                  \\
      % Llama2-70B-KPMath-Plus
      %      & 1.6M  & 48.6                  & 87.4
      % & --                    & --                    & --
      %           & --                    & --                    \\
      \hdashline % Llama3-70B
      Llama3-70B-ICL
      & --    & 44.0                  & 80.1                  & 33.5
      & 51.7                  & 10.8                  & 27.0
      & 41.2                  \\
      Llama3-70B-MetaMath
      & 0.40M & 44.9                  & 88.0                  & 31.9
      & 53.2                  & 11.6                  & 21.9
      & 41.9
      \\
      Llama3-70B-MMIQC
      & 2.3M  & 49.4                  & 89.3                  & 37.6
      & 60.4                  & 15.3                  & 23.5
      & 45.9                  \\
      Llama3-70B-VRT
      & 0.59M & 53.1                  & 90.3                  & 36.8
      & 62.8                  & 19.3                  & \textbf{28.6}
      & 48.5                  \\
      \rowcolor[rgb]{ .867, .922, .969} \model[-Llama3-70B] (Uniform)
      & 0.59M & 54.9\up{1.8}          & \textbf{90.4}\up{0.1} &
      \textbf{38.5}\up{1.7} & \textbf{64.1}\up{1.3} & 19.1\down{0.2}
      & 27.4\down{1.2}        & 49.1\up{0.6}          \\
      \rowcolor[rgb]{ .867, .922, .969} \model[-Llama3-70B]
      (Prop2Diff) & 0.59M & \textbf{56.1}\up{3.0} & 89.6\down{0.7}
      & 37.9\up{1.1}          & \textbf{64.1}\up{1.3} &
      \textbf{20.0}\up{0.7} & 28.2\down{0.4}        & \textbf{49.3}\up{0.8} \\
      \midrule
      \multicolumn{10}{c}{\textbf{7B Math-Specialized Base Model}} \\
      DeepSeekMath-7B-ICL
      & --    & 35.5                  & 64.2                  & 34.7
      & 45.2                  & 9.3                   & 23.5
      & 35.4                  \\
      DeepSeekMath-7B-Instruct
      & 0.78M & 46.9                  & 82.7                  & 37.1
      & 52.2                  & 14.2                  & 28.1
      & 43.5                  \\
      DeepSeekMath-7B-MMIQC
      & 2.3M  & 45.3                  & 79.0                  & 35.3
      & 52.9                  & 13.0                  & 23.4
      & 41.5                  \\
      DeepSeekMath-7B-KPMath-Plus
      & 1.6M  & 48.8                  & 83.9                  & --
      & --                    & --                    & --
      & --                    \\
      DeepSeekMath-7B-VRT
      & 0.59M & 53.0                  & \textbf{88.2}         &
      \textbf{41.9}         & 60.2                  & 19.1
      & 27.2                  & 48.3                  \\
      \rowcolor[rgb]{ .867, .922, .969} \model[-DSMath-7B] (Uniform)
      & 0.59M & 52.9\down{0.1}        & \textbf{88.2}         &
      40.1\down{1.8}        & 60.2                  & 21.3\up{2.2}
      & \textbf{32.5}\up{5.3} & 49.2\up{0.9}          \\
      \rowcolor[rgb]{ .867, .922, .969} \model[-DSMath-7B]
      (Prop2Diff)  & 0.59M & \textbf{53.6}\up{0.6} & 86.8\down{1.4}
      & 40.7\down{1.2}        & \textbf{61.6}\up{1.4} &
      \textbf{21.7}\up{2.6} & 32.2\up{5.0}          & \textbf{49.4}\up{1.1} \\
      \midrule
      \multicolumn{10}{c}{\textbf{7-8B General Base Model}} \\
      % Llama2-7B
      % Llama2-7B-ICL
      %      & --    & 5.8                   & 14.5
      % & 7.0                   & 9.4                   & 1.3
      %           & 10.0                  & 8.0                  \\
      % Llama2-7B-MathScale
      %      & 2.0M  & 31.1                  & 66.3
      % & 20.9                  & --                    & --
      %           & --                    & --                    &
      % --                    \\
      Llama2-7B-Xwin-Math-V1.1$^\dagger$
      & 1.4M  & 45.5                  & 84.9                  & 27.6
      & 43.0                  & 10.5                  & 15.0
      & 37.8                  \\
      \hdashline % Mistral-7B
      Mistral-7B-ICL
      & --    & 16.5                  & 45.9                  & 17.9
      & 23.5                  & 3.7                   & 14.2
      & 20.3                  \\
      Mistral-7B-WizardMath-V1.1 (RL)
      & --    & 32.3                  & 80.4                  & 23.1
      & 38.4                  & 7.7                   & 16.6
      & 33.1                  \\
      Mistral-7B-MetaMath
      & 0.40M & 29.8                  & 76.5                  & 19.3
      & 28.0                  & 5.9                   & 14.0
      & 28.9                  \\
      Mistral-7B-MMIQC
      & 2.3M  & 37.4                  & 75.4                  & 28.5
      & 38.0                  & 9.4                   & 16.2
      & 34.2                  \\
      Mistral-7B-MathScale
      & 2.0M  & 35.2                  & 74.8                  & 21.8
      & --                    & --                    & --
      & --                    \\
      Mistral-7B-KPMath-Plus
      & 1.6M  & \textbf{46.8}         & 82.1                  & --
      & --                    & --                    & --
      & --                    \\
      Mistral-7B-VRT
      & 0.59M & 38.7                  & 82.3                  & 24.2
      & 35.6                  & 8.7                   & 16.2
      & 34.3                  \\
      \rowcolor[rgb]{ .867, .922, .969} \model[-Mistral-7B] (Uniform)
      & 0.59M & 43.5\up{4.8}          & \textbf{82.6}\up{0.3} &
      26.9\up{2.7}          & 42.0\up{6.4}          & 13.2\up{4.5}
      & 16.4\up{0.2}          & 37.4\up{3.1}          \\
      \rowcolor[rgb]{ .867, .922, .969} \model[-Mistral-7B]
      (Prop2Diff) & 0.59M & 45.5\up{6.8}          & 81.1\down{1.2}
      & \textbf{29.4}\up{5.2} & \textbf{45.1}\up{9.5} &
      \textbf{14.7}\up{6.0} & \textbf{17.0}\up{0.8} & \textbf{38.8}\up{4.5} \\
      \hdashline % Llama3-8B
      Llama3-8B-ICL
      & --    & 21.2                  & 51.0                  & 19.9
      & 27.4                  & 4.2                   & 19.8
      & 23.9                  \\
      Llama3-8B-MetaMath
      & 0.40M & 32.5                  & 77.3                  & 20.6
      & 35.0                  & 5.5                   & 13.8
      & 30.8                  \\
      Llama3-8B-MMIQC
      & 2.3M  & 39.5                  & 77.6                  &
      \textbf{29.5}         & 41.0                  & 9.6
      & 16.2                  & 35.6                  \\
      Llama3-8B-VRT
      & 0.59M & 39.7                  & 81.7                  & 23.9
      & 41.7                  & 9.3                   & 14.9
      & 35.2                  \\
      \rowcolor[rgb]{ .867, .922, .969} \model[-Llama3-8B] (Uniform)
      & 0.59M & 45.3\up{5.6}          & \textbf{82.5}\up{0.8} &
      27.1\up{3.2}          & \textbf{48.2}\up{6.5} & 13.6\up{4.3}
      & 15.4\up{0.5}          & 38.7\up{3.5}          \\
      \rowcolor[rgb]{ .867, .922, .969} \model[-Llama3-8B]
      (Prop2Diff)  & 0.59M & \textbf{46.6}\up{6.9} & 81.1\down{0.6}
      & 28.8\up{4.9}          & 48.0\up{6.3}          &
      \textbf{14.5}\up{5.2} & \textbf{19.4}\up{4.5} & \textbf{39.7}\up{4.5} \\
      \bottomrule
    \end{tabular}
  }
  \caption{Main results on mathematical benchmarks. College, DM,
    Olympiad, Theorem denote the CollegeMath, DeepMind-Mathematics,
    OlympiadBench-Math, TheoremQA benchmarks respectively.
    We annotate the absolute accuracy change compared to the VRT
    baseline within the same base model.
    Bold means the best score within the respective base model.
    ICL, MetaMath, MMIQC, and VRT baselines are from our own runs,
    while other numbers are copied from the respective papers or
    reports. For WizardMath and Xwin-Math, we take the public model
    checkpoints and evaluate ourselves using their official CoT prompt.
    $^\dagger$: For Xwin-Math, we take the best public models that
    are based on Llama2~\citep{touvron2023llama}, which is not a very
    fair comparison with others.
    % \jh{maybe consider moving 70-72B in the bottom, let's see the
    % 70B results first}\jh{copy xwin-math mistral7B results from the
    % paper here}
  }
  \label{tab:main-results}
  \vspace{-15pt}
\end{table}

%% file: discussion.tex
\section{Discussion}\label{sec:discussion}

In this paper, we focus on instruction tuning for mathematical
problem solving, and discuss the impact of distribution and coverage
of training queries across different difficulties. We identify the
bias towards easy queries in vanilla rejection tuning, and propose
difficulty-aware rejection tuning, \method, as a remedy.
Based on our approach, we create and open-source the best-performing
and the most cost-effective instruction tuning datasets for
mathematical reasoning, without relying on proprietary models.
Extensive experiments across various base models and benchmarks
demonstrate the effectiveness of our approach.
% on model performance on mathematical reasoning tasks and argue that
% to obtain the best performance, enough responses to difficult
% queries are necessary, while only a few should be enough for easy
% queries. Based on this strategy, we propose \textbf{\fullmethod}
% (\method), building \texttt{\method-Math} datasets series and model
% family. Our experiments demonstrate that \model~ models show
% significant improvement in mathematical reasoning abilities
% compared to the vanilla rejection sampling baseline. Our work
% emphasizes the importance of difficult samples in training data,
% providing new insights for understanding instruction tuning for
% mathematical reasoning and building better training datasets.
\paragraph{Limitations:}
We utilize fail rate as the difficulty metric, yet it may be
sub-optimal. Other metrics such as direct
scoring~\citep{liu2024deita}, Elo ratings, or the minimum pretraining
compute to train a model that can always answer
correctly~\citep{burns2023weak} may be further explored.
% Additionally, we only focus on the difficulty aspect of synthetic
% data in this work, while other aspects like diversity may further
% improve the performance. such as increasing the diversity of the
% responses for the same query.
\model~is limited by natural language reasoning, while it is shown
that generating and executing code helps solve mathematical problems
significantly~\citep{zhou2024csv,yue2024mammoth,gou2024tora,liao2024mario,toshniwal2024openmathinstruct}
--- we think the bias in vanilla rejection sampling also exists for
code generation, and \method[] could be integrated to potentially
improve code generation as well.

%% file: appendix.tex
\appendix

\section{Comparison to Methods Based on Non-vanilla Rejection
Sampling}\label{app:comparison}

Though both ToRA and MARIO have not released their datasets and focus
on mathematical problem-solving using code
in addition to natural language, which is out of our scope and thus
not comparable, we try to implement the natural-language versions of
their data synthesis strategies, which are comparable with \sampling[].

\subsection{\sampling[] produces distributions not biased towards easy queries}

The most important difference between \sampling[] and ToRA/MARIO is
how responses are distributed across various queries --- while we
adjust the distribution either to be uniform or to favor more
difficult queries, rather than merely improving coverage,
\textbf{ToRA/MARIO mainly focus on improving coverage
  without managing the distribution explicitly, leading to datasets
that may still bias towards easy queries}.

As shown in
\autoref{tab:comparison_to_non_vanilla}, though the absolute numbers
of responses are not directly comparable between different methods,
distribution-wise we can see that ToRA/MARIO still produce fewer
responses for difficult problems than the easy ones. This especially
contrasts with \dataset[-Hard], which produces, for example, 10x more
responses for the MATH Level 5 queries than for the GSM8K queries.

As
demonstrated in \autoref{fig:cover_and_sample_cost} (Left), a high
coverage rate (VRT+Cover) alone does not guarantee superior
performance.

\input{tables/comparison_to_non_vanilla}

\subsection{\sampling[] achieves high coverage even on the hardest queries}

It is worth noting that a relatively high total coverage on MATH
training set does not mean that the hard queries are well covered.
For example, the MetaMathQA-MATH-AnsAug dataset achieves 82.8\% of
coverage on the MATH training set with evenly allocated budgets yet
still admits missing a significant portion of hard queries and biasing
towards easy queries, as analyzed in \autoref{fig:bias_pass_at_k}.

In \autoref{tab:level_wise_math_coverage} we show the coverage rate
across all the difficulty levels by different methods. The
ToRA-Corpus-16k statistics show that it only covers 68\% of the Level
5 MATH queries while DART-Math datasets cover 99.6\%.

\input{tables/level_wise_math_coverage}

\subsection{Details of Re-implementing Data Synthesis Strategies of
ToRA and MARIO}

Here we supplement more details on how we replicate the ToRA/MARIO
synthesis pipeline to conduct the analysis present in the general
author rebuttal. Below we show in the format as ``ToRA/MARIO's method -> how
we adapt similar spirits for a simpler replication'' step by step (we
  use CoT format rather than tool-integrated reasoning for a fairer
comparison with our datasets):

\paragraph{ToRA:}

\begin{enumerate}
  \item Once for each problem in MATH\&GSM8K with GPT-4,
    keeping the correct responses. -> We follow this
    with GPT-4o mini\footnote{For GPT-4o mini, we use the version of
    \texttt{gpt-4o-mini-2024-07-18} by default.}

  \item 10 trials for each problem not correctly answered
    by greedy decoding with GPT-4 and keeping up to 4 correct
    responses per problem (to form ToRA-Corpus-16k). -> We follow
    this with GPT-4o mini.

  \item Training CodeLlama models on ToRA-Corpus-16k to
    perform rejection sampling next. -> To avoid
    additional training for a fairer
    comparison, we use DeepSeekMath-7B-RL to replace the trained
    CodeLLama models here to align with DART-Math.
    \begin{enumerate}
      \item 64 trials for each problem in MATH\&GSK8K with
        CodeLlama, getting 233k distinct correct responses.
        -> We follow this with
        DeepSeekMath-7B-RL, getting 733k distinct correct responses.

      \item Correcting wrong responses by greedy decoding
        from the correct preceding portions (costing no more than
        64 trials for each problem) with CodeLLaMA-34B, getting
        69k corrected responses. -> We simplify this by re-sampling
        another up to 64 trials per problem for all the incorrect
        responses, getting 225k correct samples.

      \item Both ToRA and our adaptation: Randomly selecting up
        to 4 correct responses per problem from steps (a) and (b).
    \end{enumerate}

  \item Merge ToRA-Corpus-16k and data from step 3 to form
    the final training dataset of 69k responses. -> We exactly follow
    this to form the final
    dataset of 72k responses.

\end{enumerate}

\paragraph{MARIO:}

\begin{enumerate}
  \item Greedy decoding using GPT3.5 and GPT-4 each once for
    MATH\&GSM8K, getting two responses for each query, only correct
    ones are kept -> We follow this but use GPT-4o mini to sample two
    responses for each query.

  \item Sampling for 2 trials for each problem not correctly answered
    in step 1 using GPT-4, only correct ones are kept -> We follow this
    with GPT-4o mini.

  \item Manually correcting responses for part of the remaining
    problems, then tuning Llemma-34B on it to obtain a synthesis
    agent for next steps -> this involves human annotation and is not
    comparable to our approach. For simplicity, we adopt
    DeepSeekMath-7B-RL as the synthesis agent to align with the
    \dataset[] datasets.

  \item Sampling with 100 trials and keeping up to 4 correct
    responses per problem for the remaining unanswered MATH queries,
    achieving 93.8\% coverage on MATH -> we follow this and achieve
    91.3\% coverage on MATH.

  \item Sampling with 1 trial for new problems introduced by MetaMath
    and keeping correct ones -> this step introduces new prompts and
    would only skew the distribution of responses, if any, towards
    easy queries. We remove this step for simplicity, which would not
    affect our conclusion.
\end{enumerate}

\section{Experimental Setup}\label{app:exp_setup}

\subsection{Training Setup}\label{app:train}

We train all the models using the Transformers
library~\citep{wolf2019transformers}.

\paragraph{Sequence Packing:} To efficiently save computation wasted
by padding tokens, we employ sequence
packing~\citep{krell2021sequence}. We shuffle all samples in each
epoch before sequence packing, ensuring that the same semantic
sequences are not always in the same computation sequence.

\paragraph{Batch Size:} The computation sequence token length is set
to 4096, considering that most sequences in the training datasets are
shorter than this length. The batch size is 64, though there are
usually more than 64 samples in one batch because one computation
sequence can pack multiple semantic sequences. We disable gradient
accumulation~\citep{lin2018gradacc} by default, but when the memory
is not sufficient, we increase the number of gradient accumulation
steps and keep other settings unchanged. Specifically, we use 2
gradient accumulation steps when training Llama3-8B on 8 NVIDIA A100
GPUs under our setting.

\paragraph{Learning Rate:} We use the Adam
optimizer~\citep{zhang2018adam} with the weight decay as 0. We use a
linear warmup with a warmup step ratio of 0.03 and cosine learning
rate scheduler. The maximum learning rates are set as follows:
Mistral-7B at 1e-5, DeepSeekMath-7B and Llama3-8B at 5e-5, and
Llama3-70B at 2e-5. We determine the values by searching through
\texttt{1e-6,5e-6,1e-5,2e-5,5e-5,1e-4} according to the MATH
performance after training on MMIQC for 1 epoch.

\paragraph{\# Training Epochs:} The default number of epochs is 3.
For MMIQC, we train for 1 epoch  following \citet{liu2024mmiqc}. For
Llama3 models, we train for 1 epoch because preliminary experiments
indicate that 1 epoch consistently outperforms 3 epochs.

\paragraph{Prompt Template:} For the prompt template, we use the
format following \citet{alpaca}:

\begin{promptbox}[Prompt Template]{lightgreen}
  \texttt{Below is an instruction that describes a task. Write a
    response that appropriately completes the request.\textbackslash
    n\textbackslash n\#\#\#Instruction:\textbackslash
  n\{query\}\textbackslash n\textbackslash n\#\#\# Response:\textbackslash n}
\end{promptbox}

\paragraph{Other Details:} For efficiency, We utilize various tools /
libraries / techniques including:
\begin{itemize}
  \item the DeepSpeed distributed
    framework~\citep{rasley2020deepspeed} with
    ZeRO~\citep{rajbhandari2020zero} stage 3
  \item gradient checkpointing~\citep{chen2016gradcheckpoint}
  \item \texttt{torch.compile}~\citep{ansel2024pytorch}
  \item mixed-precision training~\citep{micikevicius2018mixed} of
    BrainFloat16~\citep{kalamkar2019bf16} and
    TensorFloat32~\citep{nvidia2020tf32}
\end{itemize}

\paragraph{Hardware:} For 7B or 8B models, we train on 8 NVIDIA A100
GPUs. For 70B models, we train on 32 NVIDIA A100 GPUs.

\paragraph{Training Time Cost}

The specific training time cost depends on too many factors to give a
precise expression, such as model architecture, model size, data
content, training algorithm implementation, hardware environment,
etc. Here we provide several data points under our setting for reference:

\begin{table}[htbp]
  \centering
  \begin{tabular}{ccccc}
    \toprule
    Dataset       &
    \begin{tabular}[c]{@{}c@{}}\# Samples\\ (k)
    \end{tabular} & Model           & Hardware     &
    \begin{tabular}[c]{@{}c@{}}Time\\ (hour/epoch)
    \end{tabular} \\
    \midrule
    \dataset[-Hard] & 585
    & DeepSeekMath-7B & 8 A100 GPUs  & 3
    \\
    \dataset[-Hard] & 585
    & Mistral-7B      & 8 A100 GPUs  & 3
    \\
    \dataset[-Hard] & 585
    & Llama3-8B       & 8 A100 GPUs  & 3
    \\
    \dataset[-Hard] & 585
    & Llama3-70B      & 32 A100 GPUs & 6
    \\
    \bottomrule
  \end{tabular}
  \caption{Examples of training time cost.}
  \label{tab:train_time}
\end{table}

\subsection{Synthesis Setup}\label{app:synthesis}

\paragraph{Generation:} We utilize the vLLM
library~\cite{kwon2023vllm}, setting the maximum number of output
tokens as 2048 and adopt top-p sampling with $p=0.95$. For
temperature $t$, we search from 0.3 to 1.8 with a step of 0.1 by
using DeepSeekMath-7B-RL to sample answer-correct responses to
queries in MATH training set. We observe the speeds to achieve
specified correct answer coverage of different temperatures and find
that, for DeepSeekMath-7B-RL, higher temperatures achieve faster, but
$t \ge 1.0$ are quite similar and $t \ge 1.7$ cause the output to be
nonsense. Besides, we find that higher temperatures produce more
diverse responses by visualizing the embedings of response from
different temperatures to the same query using
t-SNE~\citep{van2008tsne}. Finally, we set the temperature as
$t=1.6$. This choice should be fair since the temperature search is not
specifically tailored for \method[].

\paragraph{Grading:} To judge whether the answers in raw responses
are correct or not as accurately as possible, we implement an
elaborate answer extraction and judgement pipeline based on regular
expressions and SymPy~\citep{meurer2017sympy} symbolic calculation,
which is able to correctly process most mathematical objects such as
matrices (vectors), intervals, symbols besides numbers, as well as
some special texts like bool expressions, dates and times.

\paragraph{Calculating Fail Rate:} For efficiency, we merge
\sampling[-Uniform] synthesis and calculating fail rates as mentioned
in \textsection\ref{sec:dars}. Specifically, we set $k_{u} = 192$ to
synthesize our data pool, and based on all the responses sampled, we
calculate fail rate for each query as \[\text{fail rate} =
\frac{\text{\# of wrong responses}}{\text{\# of all raw responses}}\]
which would produce more accurate fail rate values but is not
necessary for general algorithm implementations.

\subsection{Evaluation Setup}

\paragraph{Generation} Like \textsection\ref{app:synthesis}, we use
the vLLM library, setting the maximum number of output tokens as 2048
and adopting top-p sampling with $p=0.95$. But we use greedy decoding
(i.e. set temperature $t=0$) for evaluation. Note that there might
still be randomness from vLLM implementation despite using greedy
decoding, so we run each evaluation in
\textsection\ref{tab:main-results} with at least 3 random seeds. When
evaluating models trained by us, we use the Alpaca~\citep{alpaca}
prompt template consistent with training as shown in
\textsection\ref{app:train}. All SFT \& RL models are evaluated with
0-shot, while all base models with few-shot in-context learning
(ICL): MATH (4-shot), GSM8K (4-shot), CollegeMath (4-shot), DeepMind
Mathematics (4-shot), OlympiadBench-Math (4-shot), TheoremQA
(5-shot). For baseline models, prompts in official implementations
are used. Specially, the CoT version of Alpaca prompt template is
used for WizardMath.

\paragraph{Grading} We utilize the same pipeline as
\textsection\ref{app:synthesis} by default, except that, for
OlympiadBench, we use the official implementation of answer
correctness judgement component by \citet{he2024olympiadbench}, which
utilizing the numerical error range information provided with query,
but keep the answer extraction component of ours, because the
official implementation fails to extract a non-negligible part of
answers, especially for base model ICL.

\section{Additional Results}\label{app:additional_results}

\subsection{Domain-wise Performance on MATH}\label{app:domain_wise_math}

We test the domain-wise performance on MATH for rejection-tuned
models based on Mistral-7B and Llama3-8B. As shown in
\autoref{tab:domain_wise_math}, both domain-wise and domain-macro-average
scores still show \method[]'s significant improvement across all domains.

\input{tables/domain_wise_math}

\subsection{\method[] achieves comparable performance with RL}

\method[] is an SFT method, which is usually not comparable with RL
method like GRPO used by DeepSeekMath-7B-RL.

However, even considering comparison with DeepSeekMath-7B-RL, we find that
sole SFT with \method[] can produce performance comparable with RL on
DeepSeekMath-7B, as shown by \autoref{tab:dsmath_dart_rl}.

\input{tables/dsmath_dart_rl}

% \layout

\input{rel_work}

%% file: tables/comparison_to_non_vanilla.tex
\begin{table}[htbp]
  \centering
  % \small
  \begin{tabular}{@{}lrrrrrrrr@{}}
    \toprule
    \multirow{2}{*}{Synthetic Dataset} & Size & RPQ in &
    \multicolumn{5}{c}{ RPQ in Level-wise MATH } & MATH \\
    \cmidrule(lr){4-8}
    & (k) & GSM8K  & 1 & 2 & 3 & 4 & 5 & Coverage \\
    \midrule
    ToRA               & 72  & 5.03  & 5.01  & 4.99  & 4.95  & 4.77  & 3.84
    & 93.4\% \\
    MARIO              & 29  & 2.02  & 2.01  & 1.98  & 1.94  & 1.89  & 1.57
    & 91.3\% \\
    \midrule
    \dataset[-Uniform] & 585 & 39.93 & 40.00 & 40.00 & 39.80 & 39.54 & 37.14
    & 99.6\% \\
    \dataset[-Hard]    & 590 & 8.49  & 14.28 & 33.52 & 54.94 & 79.59 & 107.06
    & 99.6\% \\
    \bottomrule
  \end{tabular}
  \caption{Comparison between datasets synthesized by methods based on
    non-vanilla rejection sampling. ``RPQ'' means the average number
    of responses per query. The ToRA and MARIO datasets here are
    implemented by us according to their papers' descriptions, since
  the official implementations have not been open-sourced.}
  \label{tab:comparison_to_non_vanilla}
\end{table}

%% file: tables/level_wise_math_coverage.tex
\begin{table}[htbp]
  \centering
  \begin{tabular}{lrrrrrr}
    \toprule
    MATH training set coverage & Total &
    Level 1 & Level 2 & Level 3 &
    Level 4 & Level 5 \\
    \midrule
    ToRA-Corpus-16k-MATH & 83.1\% & 97.7\% & 91.6\% & 86.5\% & 81.3\%
    & 68.0\% \\
    MetaMath-MATH-AnsAug & 82.8\% & 98.1\% & 93.6\% & 86.7\% & 76.6\%
    & 48.9\% \\
    VRT Baseline & 84.9\% & 99.6\% & 98.2\% & 95.2\% & 89.8\% & 62.9\% \\
    \midrule
    \dataset[-*] & \textbf{99.6\%} & \textbf{100.0\%} &
    \textbf{100.0\%} & \textbf{99.9\%} & \textbf{99.7\%} & \textbf{99.1\%} \\
    \bottomrule
  \end{tabular}
  \caption{MATH training set coverage rates  across all the
    difficulty levels of different synthetic datasets. The numbers of
    ToRA-Corpus-16k-MATH are from their OpenReview
    page\protect\footnotemark. The two \dataset[-*] datasets have the
    same coverage because of the ``Cover'' operation, which tries to
  ensure there is at least one correct response for each query.}
  \label{tab:level_wise_math_coverage}
\end{table}

\footnotetext{\url{https://openreview.net/forum?id=Ep0TtjVoap}}

%% file: tables/domain_wise_math.tex
\begin{table}[htbp]
  \resizebox{\textwidth}{!}{
    \begin{tabular}{lccccccccc}
      \toprule
      \multicolumn{1}{c}{\multirow{2}{*}{Model}} &
      \multicolumn{7}{c}{MATH Domains} &
      \multicolumn{2}{c}{Average} \\
      \cmidrule(lr){2-8} \cmidrule(lr){9-10}
      &
      \begin{tabular}[c]{@{}c@{}}Prob.
      \end{tabular} & Prealg. &
      \begin{tabular}[c]{@{}c@{}}Num.
      \end{tabular} &
      \begin{tabular}[c]{@{}c@{}}Interm. Alg.
      \end{tabular} & Alg. & Precalc. & Geo. & Micro & Macro \\
      \midrule
      Llama3-8B-VRT
      & 34.2 & 57.8 & 30.7 & 20.4 & 59.6 & 22.5 & 29.0 & 39.7 & 36.3 \\
      \rowcolor[rgb]{ .867, .922, .969} \model[-Llama3-8B] (Uniform)
      & 34.6 & \textbf{65.7} & 35.7 & 25.4 &
      66.6 & 29.3 & 32.4 & 45.3 &
      41.4 \\
      \rowcolor[rgb]{ .867, .922, .969} \model[-Llama3-8B] (Prop2Diff)
      & \textbf{38.8} & 62.9 & \textbf{36.8}
      & \textbf{26.1} & \textbf{67.3} &
      \textbf{32.0} & \textbf{39.9} &
      \textbf{46.6} & \textbf{43.4} \\
      \midrule
      Mistral-7B-VRT
      & 32.1 & 56.3 & 29.6 & 19.0 & 58.4 & 22.2 & 30.7 & 38.7 & 35.5 \\
      \rowcolor[rgb]{ .867, .922, .969} \model[-Mistral-7B] (Uniform)
      & 33.8 & 59.8 & 35.2
      & 24.4 & 64.1 & 28.8 & 34.2 &
      43.5 & 40.0 \\
      \rowcolor[rgb]{ .867, .922, .969} \model[-Mistral-7B] (Prop2Diff)
      & \textbf{36.1} & \textbf{61.3} & \textbf{35.4} &
      \textbf{26.0} & \textbf{65.7} &
      \textbf{31.1} & \textbf{40.5} &
      \textbf{45.5} & \textbf{42.3} \\
      \bottomrule
    \end{tabular}
  }
  \caption{MATH performance across all the domains. Macro average assigns equal
    weights to each domain, while micro average assigns equal weights to
    each query, which is the same to the whole-benchmark score. The
    full names of the domains are Counting \& Probability,
    Prealgebra, Number Theory, Intermediate Algebra, Algebra,
    Precalculus, Geometry, respectively. \textbf{Bold} means the best
  score within the respective base model.}
  \label{tab:domain_wise_math}
\end{table}

%% file: tables/dsmath_dart_rl.tex
\begin{table}[htbp]
  \centering
  \resizebox{\textwidth}{!}{
    \begin{tabular}{lccccccc}
      \toprule
      Model & MATH & GSM8K & College & DM & Olympiad & Theorem & AVG \\
      \midrule
      DeepSeekMath-7B-RL & 53.1 & \textbf{88.4} & 41.3 & 58.3 & 18.7
      & \textbf{35.9} & 49.3 \\
      \midrule
      \model[-DSMath-7B] (Uniform) & 52.9 & 88.2 & 40.1 & 60.2 &
      21.3 & 32.5 & 49.2 \\
      \model[-DSMath-7B] (Prop2Diff) & \textbf{53.6} & 86.8 & 40.7 &
      \textbf{61.6} &
      \textbf{21.7} & 32.2 & \textbf{49.4} \\
      \bottomrule
    \end{tabular}
  }
  \caption{Performance by \method[] and RL on DeepSeekMath-7B.
    College, DM, Olympiad, Theorem denote the
    CollegeMath, DeepMind-Mathematics, OlympiadBench-Math, TheoremQA
    benchmarks respectively. \textbf{Bold} means the best score within the
  respective base model.}
  \label{tab:dsmath_dart_rl}
\end{table}

%% file: rel_work.tex
\section{Related Work}

\paragraph{Rejection-Sampling-Based Data Synthesis:} Rejection
sampling \citep{neal2003slice} is a statistical approach used to generate
samples from some target distribution that is not directly accessible (e.g.,
the distribution of correct responses to all the queries).
In model training, this can be used for construting training data and
usually implemented in some form of ``sampling and filtering''. Depending on
the task, the supervision
signal for filtering can be reward models, ground-truth
answers, answer consistency, e.t.c.
\citep{bai2022hhrlhf,zelikman2022star,huang2022selfimprove,dong2023raft,gulcehre2023rest,yuan2023rft,singh2023restem}.
However, most of previous works sample the same number of candidates
for each query, regardless of the query difficulty, unconsciously
introducing a bias towards easy queries in the final training data
distribution. \method[] resolves this issue by explicitly controlling
the final distribution with adaptive budget allocation of candidate samples.

\paragraph{Data Construction for Instruction Tuning} Data have been
seen one of the most critical factor for the performance of
instruction tuning. Previous works
construct metrics for data selection and construction in diverse
ways, such as training
predictors
\citep{cao2023instructionmining,lu2024instag,liu2024deita}, prompting
LLMs \citep{chen2024alpagasus}, gradient-based metrics
\citep{xia2024less} and heuristics
\citep{li2023ifd,li2023nuggets,ning2024can}. But most of them do not
consider the final distribution of training data. \method[] focus on the
metric for difficulty and further controls the whole distribution,
providing a new perspective for data selection and construction.

%% file: checklist.tex
\newpage
\section*{NeurIPS Paper Checklist}

\begin{enumerate}

  \item {\bf Claims}
  \item[] Question: Do the main claims made in the abstract and
    introduction accurately reflect the paper's contributions and scope?
  \item[] Answer: \answerYes{} % Replace by \answerYes{},
    % \answerNo{}, or \answerNA{}.
  \item[] Justification: Main claims made in the abstract and
    introduction can accurately reflect the paper's contributions to
    instruction tuning data construction and scope of mathematical reasoning.

  \item {\bf Limitations}
  \item[] Question: Does the paper discuss the limitations of the
    work performed by the authors?
  \item[] Answer: \answerYes{} % Replace by \answerYes{},
    % \answerNo{}, or \answerNA{}.
  \item[] Justification: Limitationa are discussed in
    \textsection\ref{sec:discussion}.

  \item {\bf Theory Assumptions and Proofs}
  \item[] Question: For each theoretical result, does the paper
    provide the full set of assumptions and a complete (and correct) proof?
  \item[] Answer: \answerNA{} % Replace by \answerYes{}, \answerNo{},
    % or \answerNA{}.
  \item[] Justification: The paper does not include theoretical results.

  \item {\bf Experimental Result Reproducibility}
  \item[] Question: Does the paper fully disclose all the information
    needed to reproduce the main experimental results of the paper to
    the extent that it affects the main claims and/or conclusions of
    the paper (regardless of whether the code and data are provided or not)?
  \item[] Answer: \answerYes{}{} % Replace by \answerYes{},
    % \answerNo{}, or \answerNA{}.
  \item[] Justification: The paper fully disclose all the information
    needed to reproduce the main experimental results of the paper to
    the extent that it affects the main claims and conclusions of the
    paper. See \textsection\ref{sec:method} and
    Appendix~\ref{app:exp_setup} for details.

  \item {\bf Open access to data and code}
  \item[] Question: Does the paper provide open access to the data
    and code, with sufficient instructions to faithfully reproduce
    the main experimental results, as described in supplemental material?
  \item[] Answer: \answerYes{} % Replace by \answerYes{}, \answerNo{},
    % or \answerNA{}.
  \item[] Justification: Our datasets, models and code are publicly available at
    \url{https://github.com/hkust-nlp/dart-math}.

  \item {\bf Experimental Setting/Details}
  \item[] Question: Does the paper specify all the training and test
    details (e.g., data splits, hyperparameters, how they were
    chosen, type of optimizer, etc.) necessary to understand the results?
  \item[] Answer: \answerYes{} % Replace by \answerYes{},
    % \answerNo{}, or \answerNA{}.
  \item[] Justification: The paper specifies all the training and
    test details necessary to understand the results. See
    Appendix~\ref{app:exp_setup} for details.

  \item {\bf Experiment Statistical Significance}
  \item[] Question: Does the paper report error bars suitably and
    correctly defined or other appropriate information about the
    statistical significance of the experiments?
  \item[] Answer: \answerNo{}{} % Replace by \answerYes{},
    % \answerNo{}, or \answerNA{}.
  \item[] Justification: Experiments on LLMs are too expensive to run
    for many times.

  \item {\bf Experiments Compute Resources}
  \item[] Question: For each experiment, does the paper provide
    sufficient information on the computer resources (type of compute
    workers, memory, time of execution) needed to reproduce the experiments?
  \item[] Answer: \answerYes{} % Replace by \answerYes{},
    % \answerNo{}, or \answerNA{}.
  \item[] Justification: The paper provide sufficient information on
    the computer resources needed to reproduce the experiments. See
    Appendix~\ref{app:exp_setup} for details.

  \item {\bf Code Of Ethics}
  \item[] Question: Does the research conducted in the paper conform,
    in every respect, with the NeurIPS Code of Ethics
    \url{https://neurips.cc/public/EthicsGuidelines}?
  \item[] Answer: \answerYes{} % Replace by \answerYes{},
    % \answerNo{}, or \answerNA{}.
  \item[] Justification: The research conducted in the paper conform
    with the NeurIPS Code of Ethics in every respect.

  \item {\bf Broader Impacts}
  \item[] Question: Does the paper discuss both potential positive
    societal impacts and negative societal impacts of the work performed?
  \item[] Answer: \answerNA{} % Replace by \answerYes{}, \answerNo{},
    % or \answerNA{}.
  \item[] Justification: We use data from common public mathematical
    datasets and synthesize data only about mathematics, with little
    impact on society. We do not observe any obviously negative
    societal impacts.

  \item {\bf Safeguards}
  \item[] Question: Does the paper describe safeguards that have been
    put in place for responsible release of data or models that have
    a high risk for misuse (e.g., pretrained language models, image
    generators, or scraped datasets)?
  \item[] Answer: \answerNA{} % Replace by \answerYes{}, \answerNo{},
    % or \answerNA{}.
  \item[] Justification: We use data from common public mathematical
    datasets and synthesize data only about mathematics, with a low
    risk for misuse.

  \item {\bf Licenses for existing assets}
  \item[] Question: Are the creators or original owners of assets
    (e.g., code, data, models), used in the paper, properly credited
    and are the license and terms of use explicitly mentioned and
    properly respected?
  \item[] Answer: \answerYes{} % Replace by \answerYes{},
    % \answerNo{}, or \answerNA{}.
  \item[] Justification: The creators or original owners of assets
    used in the paper are properly credited and the license and terms
    of use are explicitly mentioned and properly respected.

  \item {\bf New Assets}
  \item[] Question: Are new assets introduced in the paper well
    documented and is the documentation provided alongside the assets?
  \item[] Answer: \answerYes{} % Replace by \answerYes{},
    % \answerNo{}, or \answerNA{}.
  \item[] Justification: The code, data and models are well
    documented and the documentation will be made publicly available
    alongside the assets following the review period.

  \item {\bf Crowdsourcing and Research with Human Subjects}
  \item[] Question: For crowdsourcing experiments and research with
    human subjects, does the paper include the full text of
    instructions given to participants and screenshots, if
    applicable, as well as details about compensation (if any)?
  \item[] Answer: \answerNA{} % Replace by \answerYes{}, \answerNo{},
    % or \answerNA{}.
  \item[] Justification: The paper does not involve crowdsourcing nor
    research with human subjects.

  \item {\bf Institutional Review Board (IRB) Approvals or Equivalent
    for Research with Human Subjects}
  \item[] Question: Does the paper describe potential risks incurred
    by study participants, whether such risks were disclosed to the
    subjects, and whether Institutional Review Board (IRB) approvals
    (or an equivalent approval/review based on the requirements of
    your country or institution) were obtained?
  \item[] Answer: \answerNA{} % Replace by \answerYes{}, \answerNo{},
    % or \answerNA{}.
  \item[] Justification: The paper does not involve crowdsourcing nor
    research with human subjects.

\end{enumerate}